\documentclass[10pt,twocolumn,letterpaper]{article}

\usepackage[pagenumbers]{cvpr}
\usepackage{amsmath}
\usepackage{amsthm}
\usepackage{amsfonts}
\usepackage{amssymb}
\usepackage{dsfont}
\usepackage{graphicx}
\usepackage{pgfplots}
\usepackage{pgfplotstable}
\usepackage{caption}
\usepackage{tabularx}    
\usepackage{multirow}    
\usepackage{booktabs} 
\usepackage{subcaption} 
\usepackage{algorithmicx}
\usepackage{algorithm}
\usepackage{algpseudocode}
\usepackage{xspace}
\usepackage{wrapfig}

\usepackage{xcolor}

\newcommand{\algrule}{\noindent\hrulefill}
\pgfplotsset{compat=1.18}
\definecolor{cvprblue}{rgb}{0.21,0.49,0.74}
\usepackage[pagebackref=true,breaklinks=true,colorlinks,citecolor=cvprblue,bookmarks=false]{hyperref}

\def\paperID{13973} 
\def\confName{CVPR}
\def\confYear{2025}

\title{Conformal Prediction and MLLM aided Uncertainty Quantification in Scene Graph Generation}

\author{Sayak Nag$^1$, Udita Ghosh$^{1}$, Calvin-Khang Ta$^2$\thanks{Work done while at UCR.}, Sarosij Bose$^{1}$, Jiachen Li$^1$, Amit K. Roy-Chowdhury$^1$\\
$^1$University of California, Riverside, USA, $^2$Dolby Laboratories, USA \\
\small{\texttt{\{snag005,ughos002,sbose007,jiachen.li\}}}\small{\texttt{@ucr.edu},}
\small{\texttt{calvin.ta@dolby.com},}
\small{\texttt{amitrc@ece.ucr.edu}
}}

\newtheorem{theorem}{Theorem}
\newtheorem{corollary}{Corollary}

\begin{document}
\maketitle
\begin{abstract}
Scene Graph Generation (SGG) aims to represent visual scenes by identifying objects and their pairwise relationships, providing a structured understanding of image content. However, inherent challenges like long-tailed class distributions and prediction variability necessitate uncertainty quantification in SGG for its practical viability. In this paper, we introduce a novel Conformal Prediction based framework, adaptive to any existing SGG method, for quantifying their predictive uncertainty by constructing well-calibrated prediction sets over their generated scene graphs. These scene graph prediction sets are designed to achieve statistically rigorous coverage guarantees under exchangeability assumptions. Additionally, to ensure the prediction sets contain the most practically interpretable scene graphs, we propose an effective MLLM-based post-processing strategy for selecting the most visually and semantically plausible scene graphs within each set. We show that our proposed approach can produce diverse possible scene graphs from an image, assess the reliability of SGG methods, and improve overall SGG performance.
\end{abstract}    
\section{Introduction}
\label{sec:intro}

Scene graphs provide a structured semantic representation of a scene depicted in an image. The nodes and edges of this semantic graph structure represent the objects and their bipartite relationships. Such semantic representation of images bridges the gap between vision and language, playing a crucial role in numerous multimodal reasoning tasks such as visual question answering~\cite{teney2017graph}, image retreival~\cite{johnson2015image}, as well as dynamic safety-critical tasks such as autonomous mobility~\cite{mcallister2017concrete} and robot path planning~\cite{pmlr-v229-ren23a,agia2022taskography}. However, despite numerous strides made in end-to-end scene graph generation (SGG), this task remains highly challenging due to inherent ambiguities arising from multiple sources, such as imprecise scene descriptions, imperfect object detection, linguistic variations (synonyms and homonyms)~\cite{Desai_2021_ICCV,chiou2021recovering, lu2016visual,nag2023unbiased}, long-tailed class distributions, and multiple viewpoint interpretations. As such, state-of-the-art SGG models tend to produce noisy scene graphs, raising the need to quantify their predictive uncertainty for reliable application to downstream tasks \cite{li2023safe}.


\begin{figure}
    \centering
    \includegraphics[width=\linewidth]{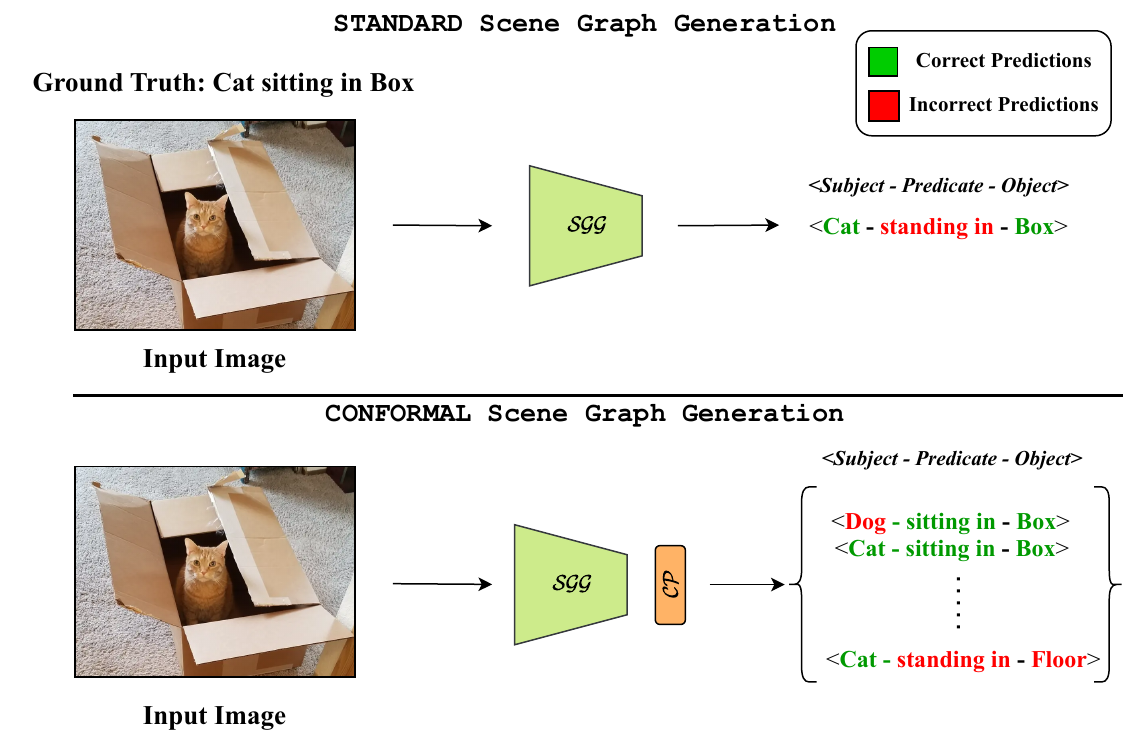}
    \caption{\textbf{Distinction between standard and conformal SGG}. The upper half shows how a standard SGG method generates a single prediction of a triplet in the image's scene graph. The lower half shows how, by adding conform prediction blocks on top of an SGG model, we can generate prediction sets for each triplet in the scene graph, which quantifies the underlying model's uncertainty and improve the chances of covering the actual ground truth.   
    }
    \label{fig:intro}
\end{figure}
While there have been works to incorporate uncertainty quantification (UQ) in SGG for performance improvement~\cite{yang2021probabilistic}, these approaches are designed specifically for their proposed model architectures. In contrast, this paper focuses on a \emph{post-hoc} distribution-free and model-agnostic UQ method with formal statistical coverage guarantees. 
Specifically, we employ \emph{Conformal Prediction} (CP)~\cite{h.papadopoulos2007, angelopoulos2023gentle, v.vovk2005,timans2024adaptive} to design a model agnostic UQ framework for state-of-the-art SGG models and provide prediction sets of an image's scene graph.
The prediction sets constructed via CP aim to \emph{provably contain the true ground-truth class with a desired probability}~\cite{angelopoulos2021uncertainty}. This enables modeling of the inherent predictive uncertainty of an underlying pre-trained SGG model, with safety assurances \cite{timans2024adaptive}. We refer to this form of SGG as \emph{Conformal SGG}, and its distinction with standard SGG is shown in Fig \ref{fig:intro}. 

Since end-to-end SGG involves detecting all objects in an image and classifying their pairwise relationships, incorporating CP entails the construction of prediction sets for the object and relationship classification. Formally, this pair of objects and their relationship is called a $<subject-predicate-object>$ triplet~\cite{krishna2017visual}, forming the fundamental block of a scene graph. Therefore, by using CP to collectively construct prediction sets for objects and predicates, we can subsequently derive a prediction set for the whole triplet. This facilitates the generation of multiple possible scene graphs for an image in the form of prediction sets with formal coverage guarantees.
The main objectives of this work are to: 1) design a CP-based post-hoc model-agnostic UQ method adaptive to any existing SGG framework, enabling the generation of multiple possible scene graphs from an image that can provide diverse possible scenarios in a downstream safety-critical task such as robot path planning~\cite{pmlr-v229-ren23a,rana2023sayplan}, 2) assess the reliability of existing state-of-the-art SGG methods by analyzing their predictive uncertainty, and 3) show that conformal SGG improves the over recall hit rate compared to standard SGG. 
As previously mentioned, long-tailed class distribution is a common problem in SGG~\cite{Desai_2021_ICCV}, as such prediction sets from standard CP, which aim to achieve aggregate coverage guarantees across the entire dataset, may fail to provide consistent coverage for each class~\cite{romano2020classification,ding2024class}, particularly those in the tail of the distribution. To address this issue, we design our SGG-specific CP framework as \emph{class-conditional} CP~\cite{romano2020classification} to ensure statistical coverage guarantees at the individual class level as opposed to the dataset level.



While CP-based prediction sets of each triplet are constructed to possibly achieve desired statistical coverage, depending on the classes captured in the individual object and predicate sets, many entries in the overall triplet prediction set may be visually uncorrelated to the scene and/or semantically implausible. This affects the interpretability of the prediction sets and has the practical drawback of large set sizes. To address this issue, we propose a post-processing filtering operation based on scene coherence and feasibility conditions using Multimodal Large Language Models (MLLM)~\cite{li2023blip}. Having been trained on extensive natural image and language pairs, these models develop an implicit understanding of real-world inter-object relationships, providing a foundational prior for selecting the most \emph{plausible} scene descriptions associated with an image. Therefore, by converting the entries in each triplet prediction set into language descriptions, we design an effective prompting strategy for in-context learning of an MLLM to achieve the desired task of selecting the most semantically plausible entries in the prediction set without compromising desired coverage guarantees. This operation improves the practicality of the prediction sets by compressing them into the most interpretable detections. We refer to the entire pipeline as \textbf{P}lausibility ensured \textbf{C}onformal \textbf{SGG}, or \textbf{PC-SGG}. 

To the best of our knowledge, \emph{this is the first work to propose a CP framework for statistically rigorous UQ for SGG, along with an MLLM-based post-processing strategy for set size minimization via plausibility assessment.} 
The main contributions are as follows:
\begin{itemize}
    \item We propose a statistically rigorous post-hoc conformal prediction framework for UQ of SGG models by generating scene graph prediction sets, which facilitates the generation of diverse scene graphs from an image.
    \item To ensure the practical viability of the prediction sets, we propose an MLLM-based post-processing strategy for filtering prediction set entries that are visually and/or semantically implausible.
    \item Using PC-SGG, we assess the reliability of numerous SGG frameworks by analyzing the empirical coverage of their prediction sets.
    \item We further demonstrate that incorporating our approach improves the recall-hit rate compared to standard SGG.  
\end{itemize}

\section{Related Works}
\label{sec:related_works}

\noindent\textbf{Scene Graph Generation.}
Scene Graph Generation (SGG) predominantly uses detection-based methods, following a two-stage framework: object detection followed by relationship detection. Visual Genome~\cite{krishna2017visual} provided foundational insights into the inherent ambiguities in human annotations of visual relationships. Key approaches include message passing structures~\cite{li2017vip}, linguistic information integration~\cite{yu2017visual}, region captioning-based methods~\cite{li2017scene}, low-dimensional vector mapping~\cite{zhang2017visual}, transformer-based networks~\cite{tang2020unbiased,li2022deep}, and feature fusion techniques~\cite{dai2017detecting}. Recent methods for relationship detection include iterative approaches~\cite{tang2020unbiased,tao2022predicate} that progressively refine relationship detector parameters, while others focus on predicate representation strategies, probabilistic modeling~\cite{yang2021probabilistic}, fine-grained predicate learning~\cite{lyu2022fine}, and hybrid-attention mechanisms~\cite{dong2022stacked}. Another direction explores sampling strategies, such as bipartite graph networks with adaptive message passing and bi-level data resampling~\cite{li2021bipartite}. However, the inherent challenges in SGG~\cite{Desai_2021_ICCV,nag2023unbiased} necessitate the need for a plug-in tool to assess the predictive uncertainty and, consequently, reliability of these approaches.

\noindent\textbf{Uncertainty in Scene Graph Generation.}
Research on uncertainty in scene graph generation (SGG) has largely addressed relationship ambiguity by exploring statistical regularities in scene graphs~\cite{zellers2018scenegraphs}, demonstrating how context influences predicate prediction. Works like~\cite{yang2018graphrcnnscenegraph} introduced an attentional graph convolutional network for implicit uncertainty handling in relationship proposals. Unbiased scene graph generation is also explored in~\cite{Tang_2020_CVPR}, addressing the uncertainty arising from long-tailed distributions in relationship detection. Methods like~\cite{Hudson2019GQAAN} included uncertainty modeling for highlighting semantic ambiguities in visual question answering through scene graphs.  However, most of these methods provide strategies for UQ specific to their model architecture and lack safety assurances regarding ground-truth coverage. In this work, we design a model-agnostic UQ method with such statistical guarantees. 

\noindent\textbf{Conformal Prediction.}
Conformal Prediction (CP) provides a framework for producing prediction sets with guaranteed coverage under the assumption of exchangeability~\cite{inbookvovk}. The method has gained significant traction in machine learning applications due to its distribution-free nature and rigorous statistical guarantees. In computer vision, CP has been successfully applied to various tasks: ~\cite{angelopoulos2021uncertainty} utilized CP for image classification uncertainty quantification, while ~\cite{pmlr-v204-andeol23a} extended it to object detection tasks. The framework has also shown promise in causal inference ~\cite{tibshirani2019conformal} and time series forecasting~\cite{NEURIPS2021_312f1ba2}. In the context of deep learning, CP has been integrated with neural networks to provide uncertainty estimates in high-stakes applications~\cite{kompa2021}. However, the application of CP to structured prediction tasks like scene graph generation remains largely unexplored, particularly when dealing with the challenge of large prediction sets while maintaining coverage guarantees.

\section{Preliminaries}
\label{sec:prelims}
\subsection{Conformal Prediction}
Assume a dataset $\mathcal{D}=\{\mathcal{D}_{tr},\mathcal{D}_{cal},\mathcal{D}_{test} \}$, comprised of a training set $\mathcal{D}_{tr}$, a calibration set $\mathcal{D}_{cal}=\{(X_i,Y_i)\}_{i=1}^{n} \sim P_{XY}$, and a test set $\mathcal{D}_{test}=\{(X_j,Y_j)\}_{j=n+1}^{n+n_t}\mspace{-5mu} \sim\mspace{-5mu} P_{XY}$. Now given a pre-trained model, $f$, for an unseen test sample $X_{n+1}$, generalized CP ~\cite{h.papadopoulos2007,rf.barber2021} entails the construction of a prediction set $\hat{\mathcal{C}}(X_{n+1}) = \text{SET}(f(X_{n+1}),\mathcal{A}, \mathcal{D}_{cal})$, where $\text{SET}(\cdot)$ is a function that constructs sets from the outputs $f(X_{n+1})$ based on $\mathcal{D}_{cal}$ and a \emph{nonconformity measure} $\mathcal{A}$. Under the assumption of exchangeability (relaxed i.i.d)~\cite{timans2024adaptive}, i.e. $\mathcal{D}_{cal}\cup(X_{n+1},Y_{n+1})$, $\hat{\mathcal{C}}(X_{n+1})$ attains the following \emph{marginal} probabilistic guarantee,
\begin{equation}
    P(Y_{n+1} \in \hat{\mathcal{C}}(X_{n+1})) \geq 1 - \alpha
    \label{eq:marginal_cp}
\end{equation}
for some tolerated \emph{miscoverage} rate $\alpha$~\cite{g.shafer2008}.
This marginal guarantee provided by generalized CP holds in average across $\mathcal{D}_{test}$ provided $\mathcal{D}_{cal}$ and $\mathcal{D}_{test}$ came from a fixed distribution~\cite{timans2024adaptive}. 
 In contrast, if granular guarantees are required (for example, at class level or cluster level), \emph{conditionally} valid coverage is necessary~\cite{y.romano2020a, m.sesia2021,c.jung2022}.
In this work, we are particularly interested in \emph{class-conditional} coverage~\cite{m.cauchois2021, v.vovk2005, m.sadinle2019a}, yielding the following statistical guarantee, for a test sample $X_{n+1}$,
\begin{equation}
    P(Y_{n+1} \in \hat{\mathcal{C}}(X_{n+1}) \mid Y_{n+1} = y) \geq 1 - \alpha_y \quad \forall y \in \mathcal{Y}
    \label{eq:class_conditional_cp}
\end{equation}
where, $\mathcal{Y}=\{1, \dots, K\}$ are distinct class-labels for a multi-class classification problem, and $\{\alpha_y\}_{y\in\mathcal{Y}}$ are classwise miscoverage rates. Eq. \ref{eq:class_conditional_cp} implies Eq. \ref{eq:marginal_cp} and provides a stronger guarantee as it controls coverage levels within each class~\cite{romano2020classification,ding2024class,timans2024adaptive}.

\subsection{Scene Graph Generation}
SGG entails obtaining a graph-structured representation, $G_t = \{S,R,O\}$, of an image $I$, describing the atomic inter-object interactions of the scene in $I$. It is achieved by detecting and subsequently classifying all the objects in a scene and then classifying their pairwise relationships. 
Here $S = \{s_{1},s_{2},...,s_{N_o} \}$ and $O = \{o_{1},o_{2},...,o_{N_o} \}$ are respectively the subject and object list both of which map to the same list of $N_o$ objects present in $I$, and $R = \{r_{1},r_{2},...,r_{N_r} \}$ is the list of bipartite object relationships called predicates. Thus, $G_t$ is obtained by combinatorially arranging subject-object pairs into triplets (\scalebox{0.95}{ $<subject-predicate-object>$}), $t_{ij} = (s_{i},r_{k},o_j)$. We refer to the set of object and predicate classes as $\mathcal{Y}_{o} = \{y_{1}^o,y_{2}^o,...,y_{K_o}^o \} $ and $\mathcal{Y}_{r} = \{y_{1}^r,y_{2}^r,...,y_{K_r}^r \} $ respectively. 
\section{Method}
\begin{figure*}
    \centering
    \includegraphics[width=1\linewidth]{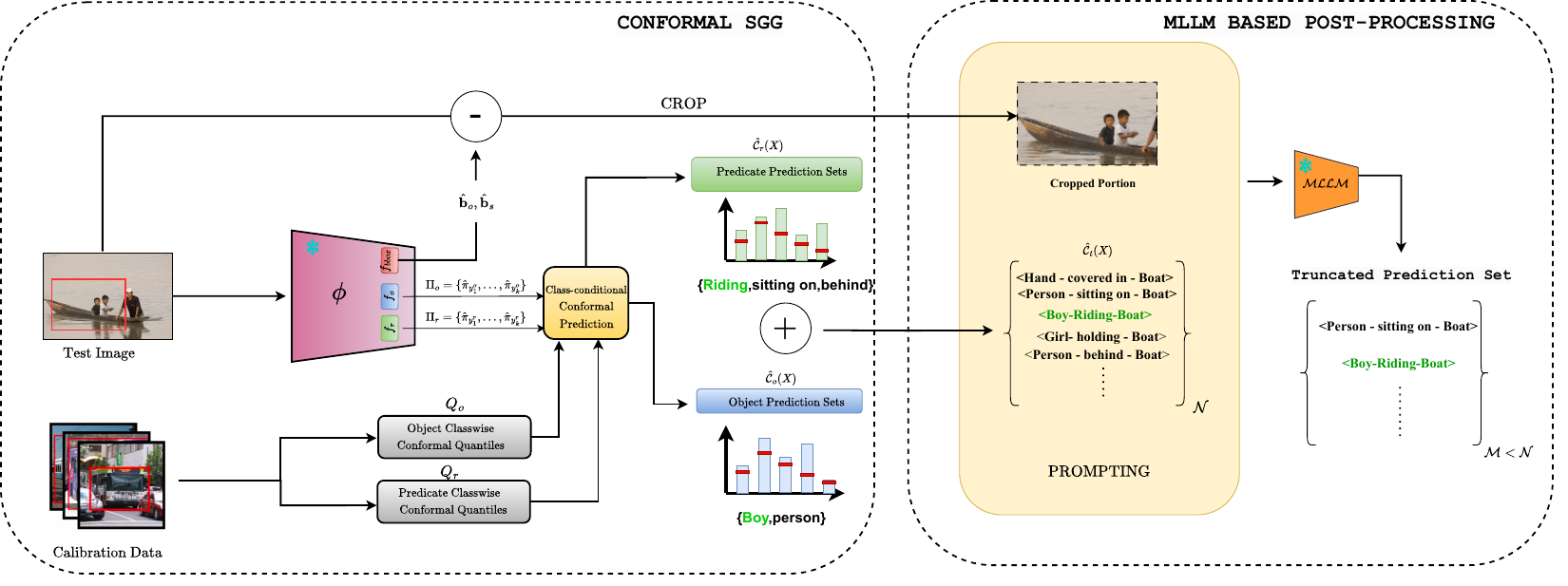}
    \caption{\textbf{Overview of PC-SGG Pipeline.} For each \emph{Test Image}, a pre-trained SGG model, $\phi$, is used to obtain object bounding boxes $\hat{\mathbf{b}}_o$ (using $f_{bbox}$), object classification probabilities $\Pi_o$ (using $f_{o}$), and the probabilities of their pairwise predicates classifications which include the classification $\Pi_r$ (using $f_{r}$). Using quantiles ($Q_o, Q_r$) derived from a \emph{Calibration Data}, we construct class-conditional conformal sets for both objects ($\hat{\mathcal{C}}_o(X)$) and predicates ($\hat{\mathcal{C}}_r(X)$). These conformal sets are then combinatorially combined ($\oplus$) to generate a \emph{Triplet Prediction Set}, $\hat{\mathcal{C}}_t(X)$. To assess the plausibility of each entry of $\hat{\mathcal{C}}_t(X)$, we leverage an MLLM-based post-processing unit. The entries of $\hat{\mathcal{C}}_t(X)$ are converted into textual descriptions, which, along with the cropped portion of the test image defined for the triplet set (cropped using the union bounding box of the triplet's object pairs) is converted into an input prompt for the MLLM to process and predict the \textit{truncated prediction set} of the most plausible triplets as a next token prediction problem. }
    \vspace{-1em}
    \label{fig:pipeline}
\end{figure*}

The overview of PC-SGG is shown in Figure \ref{fig:pipeline}. Given a pre-trained SGG model $\phi$, we first apply class-conditional CP to the object and predicate classifier heads ($f_o,f_r$) and obtain prediction sets for the objects and predicates, respectively. The two prediction sets are then combined to obtain a prediction set for each detected triplet. Each of these triplet prediction sets is then converted into sets of language descriptions to be assessed by an MLLM-based post-processing unit. Leveraging a question-answer strategy, the MLLM assesses each of these descriptions and compresses each triplet prediction set into the set of most plausible triplets. In the subsequent sections, we describe in detail the conformal prediction setup for SGG and the MLLM-based post-processing method.
%
\subsection{Conformal Scene Graph Generation}
For any test sample $X_{n+1}$ we denote the object and predicate prediction sets as $\hat{\mathcal{C}}_o(X_{n+1})$ and  $\hat{\mathcal{C}}_r(X_{n+1})$ respectively. To prevent confusion,  it must be noted that a sample, $X$, for the case of the objects, refers to the region of interest (RoI)~\cite{ren2015faster}, and for the case of predicates (and triplets), refers to a pair of detected objects which, among other information, includes the RoI associated with union box of the pair~\cite{zellers2018neural,jung2023devil,li2021bipartite}. 
Inspired by ~\cite{m.sadinle2019a}, we design the overall CP framework as a class-conditional prediction set classifier. The overall algorithm is shown in Algorithm 1 (refer to supplementary A.1).

\noindent\textbf{Conformal Calibration.}\label{sec:ConSGG}
In order to achieve the nominal class-conditional coverage of Eq. \ref{eq:class_conditional_cp}, well-calibrated prediction sets need to be constructed to measure how well an unseen test sample \emph{conforms} to the training distribution. As such, using an appropriate non-conformity measure, $\mathcal{A}$, classwise non-conformity scores are obtained on $\mathcal{D}_{cal}$, which are then used to obtain class-wise conformal quantile values, $\hat{q}_y$, that act as the necessary thresholds for assessing the conformity of any test sample. In our setup, we define the non-conformity measure as $\mathcal{A}(f(X),y) = 1-\hat{\pi}_y$, where $\hat{\pi}_y$ is the softmax probability output for class $y$ such that $f(X)=\{\hat{\pi}_y\}_{y\in \mathcal{Y}}$.

As shown in Algorithm 1, calibration proceeds by first greedily matching each pair of predicted objects to ground truth as follows,
\begin{equation}
\scalebox{0.95}{$
    \begin{aligned}
       & match(\hat{\mathbf{b}}_s,\mathbf{b}_s^j,\hat{\mathbf{b}}_o,\mathbf{b}_o^j) = \underset{j}{\text{argmax}}(\frac{1}{2}(\Gamma(\hat{\mathbf{b}}_s,\mathbf{b}_s^j)+\Gamma(\hat{\mathbf{b}}_o,\mathbf{b}_o^j)) \\
       & \text{s.t.} \ \ \ \Gamma(\hat{\mathbf{b}}_s,\mathbf{b}_s^j) \geq 0.5 \ \land \ \Gamma(\hat{\mathbf{b}}_o,\mathbf{b}_o^j) \geq 0.5
    \end{aligned}
    $}
\end{equation}
where $\Gamma$ is the Intersection over Union (IoU) function, $\hat{\mathbf{b}}_s,\hat{\mathbf{b}}_o\in \mathbb{R}^4$ are the predicted subject-object pair's bounding boxes, $\mathbf{b}_s^j,\mathbf{b}_o^j \in \mathbb{R}^4$ are the bounding boxes of the $j^{th}$ ground-truth subject-object pair. Using the matched ground truth classes (objects and predicate), the class-specific non-conformity score is computed for both the objects and the predicate. Therefore, for the $k^{th}$ object class, the aggregated list of non-conformity scores is obtained as $S_{y_k^o} = \{\mathcal{A}(f_o(X_i),Y_k^o)\}_{i=1}^{n_{y_k^o}} = \{s_i^o\}_{i=1}^{n_{y_k^o}}$, where $y_k^o \in \mathcal{Y}_o$. Similarly for the $k^{th}$ predicate class we get $S_{y_k^r} = \{\mathcal{A}(f_o(X_i),Y_k^r)\}_{i=1}^{n_{y_k^r}} = \{s_i^r\}_{i=1}^{n_{y_k^r}}$ where $y_k^r \in \mathcal{Y}_r$. Finally, the class-specific conformal quantiles are obtained as follows,
\begin{equation}
\scalebox{0.84}{$
    \begin{aligned}
        \hat{q}_{y_k^o} = \left\lceil (n_{y_k^o} + 1)(1 - \alpha_{y_k^o}) / n_{y_k^o} \right\rceil\text{-th empirical quantile of } S_{y_k^o}. \\
\hat{q}_{y_k^r} = \left\lceil (n_{y_k^r} + 1)(1 - \alpha_{y_k^r}) / n_{y_k^r} \right\rceil\text{-th empirical quantile of } S_{y_k^r}.
    \end{aligned}
    $}
    \label{eq:conformal_quantiles}
\end{equation}
The list of class-wise conformal quantiles for the objects and predicates is denoted as $Q_{o}=\{\hat{q}_{y_k^o}\}_{y_k^o \in \mathcal{Y}_o}$ and  $Q_{r}=\{\hat{q}_{y_k^r}\}_{y_k^r \in \mathcal{Y}_r}$,  respectively.
For simplicity, we set $\alpha_{y_k^o}=\alpha_o \ \forall \ y_k^o \in \mathcal{Y}_o $ and $\alpha_{y_k^r}=\alpha_r \ \forall \ y_k^r \in \mathcal{Y}_r $.
\\
\noindent\textbf{Conformal Inference.}
During inference, utilizing $Q_{o}$ and $Q_r$, $\hat{\mathcal{C}}_o(X_{n+1})$ and  $\hat{\mathcal{C}}_r(X_{n+1})$ are obtained as follows,
\begin{equation}
    \begin{aligned}
        \hat{\mathcal{C}}_o(X_{n+1}) = \{y_k^o \in \mathcal{Y}_o : \hat{\pi}_{y_{k}^o} \geq 1- \hat{q}_{y_k^o}\} \\
        \hat{\mathcal{C}}_r(X_{n+1}) = \{y_k^r \in \mathcal{Y}_r : \hat{\pi}_{y_{k}^r} \geq 1- \hat{q}_{y_k^r}\}
    \end{aligned}
    \label{eq:conformal_inference}
\end{equation}
where the meaning of an unseen test sample $X_{n+1}$, for the object and predicate cases is defined earlier.
As shown in Fig \ref{fig:pipeline}, $\hat{\mathcal{C}}_o(X_{n+1})$ and $\hat{\mathcal{C}}_r(X_{n+1})$ are combinatorially aggregated to obtain the prediction set for the entire triplet $\hat{\mathcal{C}}_t(X_{n+1})$. 
The strength of the coverage guarantees for the object and predicate prediction sets control the overall coverage guarantee for the triplet. 
\begin{theorem}
    Given the ground truth class of the $k^{th}$ triplet is denoted as $y_k^t=[y_{k}^s,y_{k}^r,y_{k}^o] \in \mathbb{R}^3$ where $y_{k}^s,y_{k}^o \in \mathcal{Y}_o$ and $y_{k}^r\in \mathcal{Y}_r$, the triplet coverage guarantee is given as $P(y_k^t \in \hat{\mathcal{C}}_t(X_{n+1}^r)) = P(Y_{n+1}^o \in \hat{\mathcal{C}}_o(X_{n+1}^o) \mid Y_{n+1}^o = y_i^o) \cdot P(Y_{n+1}^r \in \hat{\mathcal{C}}_o(X_{n+1}^r) \mid Y_{n+1}^r = y_m^r) \ \forall \ y_{k}^s\in \mathcal{Y}_o, y_{k}^o \in \mathcal{Y}_o, y_{k}^r \in \mathcal{Y}_r$.
    \label{eq:triplet_coverage}
\end{theorem}
\noindent Please see Proof in supplementary. 
\setcounter{corollary}{\value{corollary}} 
\begin{corollary}
    Following Theorem 1, $P(y_k^t \in \hat{\mathcal{C}}_t(X_{n+1}^r)) \geq (1-\alpha_o)(1-\alpha_r), \ \forall \ y_{k}^s\in \mathcal{Y}_o, y_{k}^o \in \mathcal{Y}_o, y_{k}^r \in \mathcal{Y}_r$.
\end{corollary}

To ensure a nominal coverage of each triplet prediction set to be approximately close to $90\%$, we choose $\alpha_o=0.05$ and $\alpha_r=0.1$ in our experiments.
\begin{figure}
    \centering
    \includegraphics[width=1\linewidth]{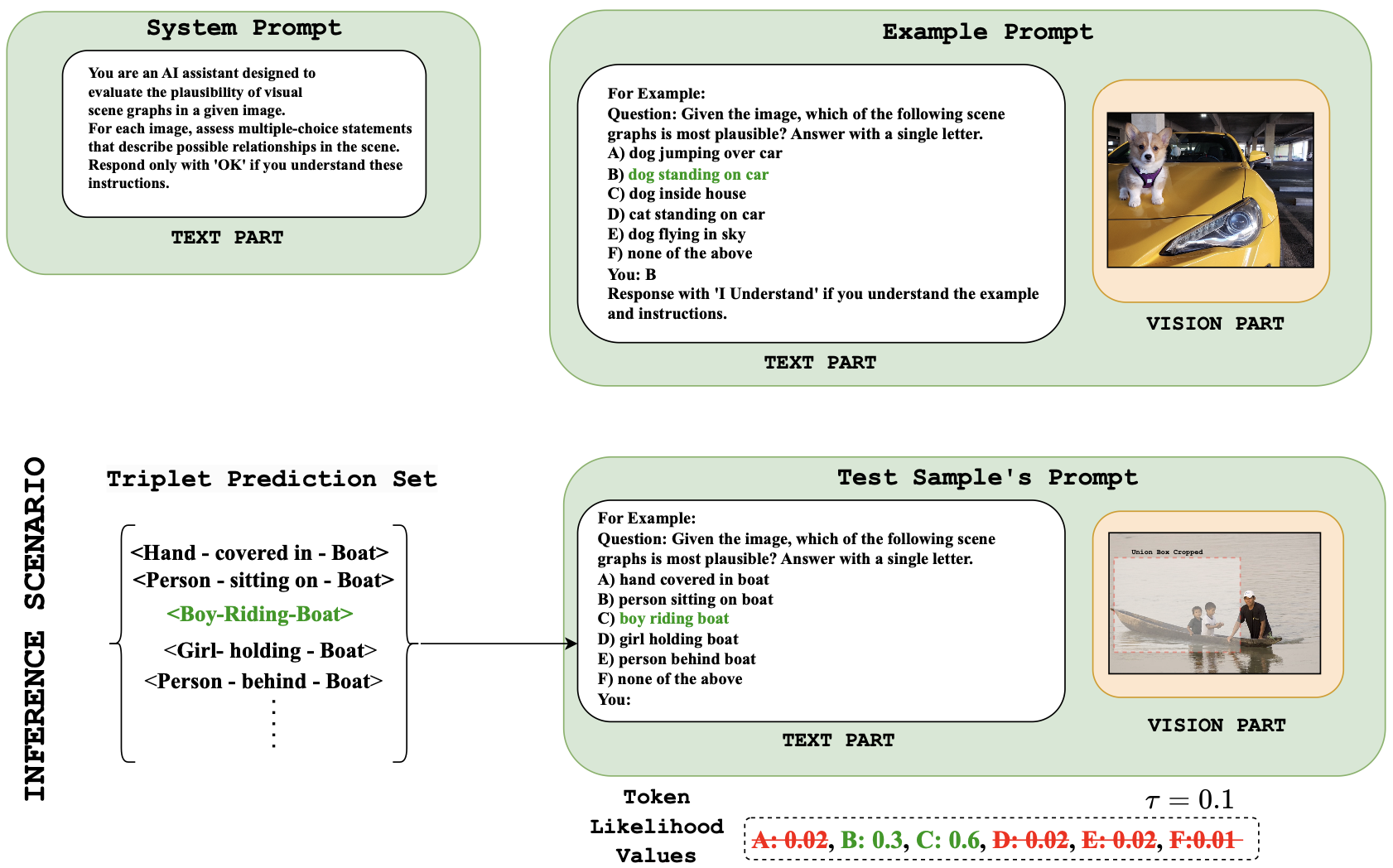}
    \caption{\textbf{Prompting strategy for plausibility assessment.} First, a \textit{system prompt} outlines the task for the MLLM. Then, an \textit{example prompt} is created with a randomly sampled image from the calibration set and a hand-crafted text description, framing plausibility assessment as an MCQA problem. During inference, entries from a test image's triplet prediction set are processed in groups of $5$, for designing the MCQ text prompts similar to the example. The vision part of the prompt is the cropped portion of the test image linked to the detected triplet. The MLLM’s token likelihoods are thresholded by $\tau$ to identify the most plausible choices for the scene. The ground-truth triplet is highlighted in green for both the example and inference scenarios.}
    \vspace{-1em}
    \label{fig:prompting}
\end{figure}

\subsection{MLLM Guided Plausibility Assessment}\label{sec:MLLM}
While the triplet prediction sets constructed using conformal SGG have nominal coverage guarantees, in practice, many entries in a triplet prediction set may not be plausible. For example, for the image shown under the \emph{Inference Scenario} in Fig \ref{fig:prompting}, its triplet prediction set has entries like ``\texttt{hand covered in boat}, \texttt{girl holding boat}". The first description of the triplet is implausible from a practical linguistic/semantic perspective, while the second one does not correlate to the actual scene in the image and is visually implausible. This causes the overall prediction set of the triplet to be bloated with impractical and uninterpretable scene descriptions. We propose a plausibility checking unit which addresses this issue by leveraging the foundational prior of an MLLM in associating the right scene description to the scene at hand. The overview of this unit's functioning is shown in Fig \ref{fig:prompting}.

\noindent\textbf{Plausibility as multiple-choice Q\&A.} We craft an effective prompting strategy that query's an MLLM regarding the plausibility of an entry in a triplet prediction set. For any detected triplet in a test image, we have the predicted bounding boxes for its subject, $\hat{\mathbf{b}}_s$, and object, $\hat{\mathbf{b}}_o$, as well as, the prediction set $\hat{\mathcal{C}}_t(X)$ obtained via the conformal prediction. First, we convert the entries of $\hat{\mathcal{C}}_t(X)$ into language descriptions, then sequentially sample $5$ language descriptions to create a multiple choice question (MCQ), forming the \emph{text part} of the prompt to the MLLM (Fig \ref{fig:prompting}). The \textit{vision part} of the prompt is composed of the cropped-out portion of the test image defined by the union box $\hat{\mathbf{b}}_s \bigcup \hat{\mathbf{b}}_o$. The MLLM is then tasked with answering the MCQ, thereby choosing the most plausible scene description among the $5$ choices. This boils down the next-token prediction of the MLLM into a single-token prediction, with the token space being restricted to the number of choices in the MCQ~\cite{srivastava2023beyond,hendrycksmeasuring}. In practice, we add a `no valid option' choice to MCQ as shown in Fig \ref{fig:prompting}. In this way, we task the MLLM to eliminate the implausible entries in each $\hat{\mathcal{C}}_t(X_{n+1})$ and compress it into a set of practically interpretable scene descriptions. 

In order to ensure the MLLM adheres to the task at hand, we leverage a one-shot in-context learning setup~\cite{brown2020language} and provide adequate task-specific information to the MLLM in the form of an instructional \textit{system prompt} and a toy scenario of the task at hand in the form of an \textit{example prompt} (Fig \ref{fig:prompting}). The \textit{example prompt} contains a randomly sampled calibration set image along with a hand-crafted toy MCQA, containing the ground-truth triplet as one of the options.

\noindent\textbf{Extraction of plausible scene descriptions.} The \textit{system prompt} and the \textit{example prompt} are passed as input to the MLLM just once, while the \emph{Inference Scenario} is run iteratively over all triplet prediction sets across all test images. The prompt of each iterative sample is essentially the cropped portion of a test image associated with the detected triplet and $5$ descriptions from that triplet's prediction set used to make the MCQ. Since the token space is restricted to the number of choices in the MCQ plus the `no valid option' choice, we obtain likelihood values for a total of $6$ tokens from the MLLM. We then threshold these likelihood values using a token threshold, $\tau$, (Fig \ref{fig:prompting}), to choose the most plausible options and aggregate them as the final truncated triplet prediction set. If the `no valid option' has the highest likelihood value, the group of $5$ prediction set entries is wholly disregarded. The choice of $\tau$ is very important for this operation. This is because the token likelihood values from the MLLM may be high for certain choices, which may be plausible w.r.t. the scene but may not be the actual ground truth triplet. So, choosing a high value for $\tau$ may result in the omission of the ground truth (provided it is covered by the original prediction set) from the truncated prediction set, consequently affecting nominal coverage. 
\section{Experiments}

\subsection{Dataset and Splits}
We perform experiments on the benchmark image scene graph generation dataset Visual Genome (VG)~\cite{krishna2017visual}, composed of $108$k images. To filter out the predicate classes with an extremely low number of samples we adopt the commonly used VG split called VG$150$ ~\cite{li2017scene,zellers2018neural}, composed of the most frequent $150$ object and $50$ predicate classes.

\noindent\textbf{Training and Calibration Sets.} 
The calibration set is carved out from the $75$k images in the original VG$150$ training and validation sets.
The calibration set is designed such that its images cover at most $10\%$ of the total number of samples in each of the object and predicate classes, ensuring that a minimum of $2$ samples from any class is included in the calibration set. The final calibration set contains $7174$ images. The remaining images are used for training.

\subsection{Implementation and Evaluation Metrics}
We incorporate PC-SGG with existing SOTA SGG methods and quantify their uncertainty for end-to-end SGG, formally called SGDET (joint detection and classification of objects and pairwise predicates). For all object classes, we set, $\alpha_o$ to $0.05$, and for all predicate classes we set $\alpha_r$ to $0.1$, This maintains the combined marginal coverage probability, $(1-\alpha_o)(1-\alpha_r)$, is close to $90\%$. The value of $\tau$ is set to $0.1$. All experiments were performed using $4$ NVIDIA RTX-$3090$.

\noindent\textbf{SGG Models.} We implement PC-SGG over the following $5$ SOTA SGG frameworks, MOTIFS~\cite{zellers2018neural}, MOTIFw/DEBIAISNG (MOTIFS-D)~\cite{cui2019class}, VCTREE~\cite{tang2019learning}, BGNN~\cite{li2021bipartite}, and SQUAT~\cite{jung2023devil}. For all models, Faster R-CNN~\cite{ren2015faster} with ResNeXt-101-FPN backbone is used as the object detector. The hyperparameters specific to each model are kept the same as reported in their respective papers. Since the VG150 training set is different from the usual one, we retrain the object detector and all the models from scratch on the modified training set for our experiments.

\noindent\textbf{MLLM.} We use BLIP-2~\cite{li2023blip} with FLAN-T5-XL~\cite{JMLR:v25:23-0870} as our MLLM. Due to computational resource constraints, we restrict our experiments to the $8$-bit quantized model. Additional implementation details are in the supplementary.

\noindent\textbf{Conformal Prediction Metrics.} The following CP metrics~\cite{angelopoulos2023conformal} are used to gauge the uncertainty associated with both object and predicate classification of each SGG model,
\begin{enumerate}
    \item \emph{Coverage Validity}: Computes empirical coverage of the CP procedure and checks its deviation from overall nominal coverage guarantees~\cite{timans2024adaptive,vovk2012conditional},
    \begin{equation}
        Cov = 100\times\frac{1}{n_t} \sum\limits_{j=n+1}^{n+n_t} \mathds{1}[Y_j \in \hat{\mathcal{C}}(X_j)]
    \end{equation}
where, \(\mathds{1}[\cdot]\) is the indicator function. 
    \item \emph{Average Class
Coverage Gap}: The following metric is specific to class-conditional coverage as it measures class-wise deviation from its desired coverage level \cite{ding2024class},
\begin{align}
    CovGap = 100\times\frac{1}{|\mathcal{Y}|}\sum\limits_{y \in \mathcal{Y}}||\hat{c}_{y} - (1-\alpha_y)||_1 \\
    \hat{c}_{y} = \frac{1}{|\mathcal{D}^{y}|}\sum\limits_{j\in \mathcal{D}^{y}}\mathds{1}[Y_i \in \hat{\mathcal{C}}(X_i)] \nonumber
\end{align}
where, \(\mathcal{D}^{y} \subset \mathcal{D}^{test}\) are all samples belonging to class $y$, $||\cdot||_1$ is the $\ell_1$ distance, $\alpha_y$ is the class specific marginal error rate. For all object classes $\alpha_y=\alpha_o$ and for all predicate classes $\alpha_y=\alpha_r$. Individually for the object and predicate classes 

\item \emph{Average Set Size}: The following computes sharpness of the computed prediction sets,
\begin{equation}
    AvgSize = \frac{1}{n_t}\sum\limits_{j=n+1}^{n+n_t}|\mathcal{C}(X_j)|
\end{equation}

\item \emph{Triplet Coverage Validity}. The $Cov$ metric is used to quantify CP performance for the object and predicate classifications individually. To assess the validity of the prediction sets associated of the whole triplet, we introduce the following metric to gauge the empirical coverage of the whole triplet prediction set, 
\begin{equation}
\resizebox{0.9\linewidth}{!}{ 
$
\begin{aligned}
    Cov_T &= 100 \times \frac{1}{n_t} 
    \sum\limits_{j=n+1}^{n+n_t} m_j^s \land m_j^r \land m_j^o \\
    &m_j^s = \mathds{1}[Y_j^s \in \hat{\mathcal{C}}_o(X_j^s)], \quad
    m_j^r = \mathds{1}[Y_j^r \in \hat{\mathcal{C}}_r(X_j^r)], \\
    &m_j^o = \mathds{1}[Y_j^o \in \hat{\mathcal{C}}_o(X_j^o)]
\end{aligned}
$
}
\end{equation}
where, $\hat{\mathcal{C}}_o(X_j^s)$, $\hat{\mathcal{C}}_r(X_j^r)$, and $\hat{\mathcal{C}}_o(X_j^o)$ are the prediction sets of the subject, predicate, and object in the triplet, $Y_j^s,Y_j^o \in \mathcal{Y}_o$, and $Y_j^r \in \mathcal{Y}_r$, and $n_t$ here runs over all the ground-truth triplet samples.
\end{enumerate}

\begin{table}[t]
    \centering
    \caption{Comparison of CP metrics of different SGG methods with PC-SGG. The $Cov_T$ values shown here are computed over the truncated triplet set obtained after MLLM-based post-processing. The best results are highlighted in bold.}
    \resizebox{1\linewidth}{!}{%
    \begin{tabular}{l c c c c c c c}
        \toprule
        \multirow{2}{*}{\textbf{Method}} & \multicolumn{3}{c}{\textbf{Objects}} & \multicolumn{3}{c}{\textbf{Predicates}} & \multicolumn{1}{c}{\textbf{Triplets}} \\
        \cmidrule(lr){2-4} \cmidrule(lr){5-7} \cmidrule(lr){8-8}
        & \textbf{$Cov\uparrow$} & 
        \textbf{$CovGap\downarrow$} & \textbf{$AvgSize\downarrow$} & \textbf{$Cov\uparrow$} & 
        \textbf{$CovGap\downarrow$} &\textbf{$AvgSize\downarrow$} & \textbf{$Cov_T\uparrow$} \\
        \midrule
        
        MOTIFS~\cite{zellers2018neural}   & 88.94 & 5.8 & 4.87 & 84.11 & 6.2 & 16.09 & 74.97 \\
        MOTIFS-D~\cite{cui2019class} & 88.94 & 5.8&4.87 & 86.67 &5.9 & 16.81 & 76.67 \\
        VCTREE~\cite{tang2019learning}   & 89.38 & 5.7&\textbf{4.23} & 88.61&5.9 & 16.41 & 80.06 \\
        SQUAT~\cite{jung2023devil}    & 90.26 &4.9 & 4.48 & \textbf{90.25} & \textbf{4.6} & \textbf{14.48} & 80.25 \\
        BGNN~\cite{li2021bipartite}     & \textbf{90.35} & \textbf{4.8} & 4.48 & 89.68 & 5.2 & 16.23 & \textbf{80.45} \\
        \bottomrule
    \end{tabular}
    }
    \label{tab:coverage_comparison}
    \vspace{-1em}
\end{table}
\begin{table}[t]
    \centering
        \caption{Comparison of empirical coverage and average set size of triplet prediction sets with and without MLLM-based post-processing.}
        \resizebox{1\linewidth}{!}{%
    \begin{tabular}{lcccc}
        \hline
        \multicolumn{1}{l}{\multirow{2}{*}{\textbf{Method}}} & 
        \multicolumn{2}{c}{\textbf{w/o MLLM Plausibility Assessment}} & 
        \multicolumn{2}{c}{\textbf{w/ MLLM Plausibility Assessment}} \\ 
        \cmidrule(lr){2-3} \cmidrule(lr){4-5} 
        & $Cov_T \uparrow$ & $AvgSize \downarrow$ & $Cov_T \uparrow$ & $AvgSize \downarrow$ \\ 
        \hline
        MOTIFS~\cite{zellers2018neural}   & 74.97 & 866.09 & 74.97 & 403.21 \\
        MOTIFS-D~\cite{cui2019class} & 76.93 & 893.21 & 76.67 & 411.58 \\
        VCTREE~\cite{tang2019learning}   & 80.06 & 818.76 & 80.06 & 389.24 \\
        SQUAT~\cite{jung2023devil}    & 80.43 & 816.68 & 80.25 & 398.67 \\
        BGNN~\cite{li2021bipartite}    & 80.45 & 971.69 & 80.45 & 464.11 \\
        \hline
    \end{tabular}
}
\vspace{-1em}
\label{tab:llm_results}
\end{table}

\noindent\textbf{SGG Metrics.}\label{subsec:sgg_metrics}
The common metrics used for quantifying SGG performance are Recall@\textit{K} (R@\textit{K}) and mean-Recall@\textit{K}(mR@\textit{K}). However, since PC-SGG provides prediction sets for each detected triplet, we modify the computation of the recall hit rate. Unlike R@\textit{K} where the equality of the subject, predicate, and object classes in a detected triplet is checked to that of a matched ground truth, we check if the matched ground truth \emph{is present in the triplet prediction set} obtained from PC-SGG. We refer to these metrics, as coverage-Recall@\textit{K} (cR@\textit{K}) and coverage-mean-Recall@\textit{K} (cmR@\textit{K}) which are specific to Conformal SGG and can be considered equivalent to standard SGG metrics R@\textit{K} and mR@\textit{K}.

\subsection{Comparative Uncertainty Quantification of SGG Methods}
The empirical coverage of the object, predicate, and triplet prediction sets, obtained by PC-SGG for each SGG method is shown in Table \ref{tab:coverage_comparison}. For all the methods, the object prediction sets fail to achieve the class-conditional coverage guarantee of $95\%$ (Eq \ref{eq:class_conditional_cp}). In comparison, the predicate prediction sets of VCTREE, SQUAT, and BGNN come close to the coverage guarantee of $90\%$. However, for all the methods, $AvgSize$ of the predicate prediction sets is significantly higher than that of the objects. This shows that predicate classification uncertainty is significantly higher than object classification. As such, the predicate prediction sets are larger to meet coverage guarantees. Overall, SQUAT and BGNN achieve the best empirical coverage for their triplet prediction sets, with SQUAT having the smallest $AvgSize$ for its predicate prediction sets. From the $CovGap$ values, it can also be observed that the predicate prediction sets of SQUAT achieve the lowest class-wise deviation from the class-conditional coverage guarantee. For the object prediction sets, BGNN has lowest $CovGap$. The empirical coverage values of the triplet prediction set, $Cov_T$, also verify Theorem \ref{eq:triplet_coverage} (check supplementary) since it is approximately close to the product of the object and predicate prediction sets' $Cov$ values. However, the $Cov_T$ values of none of the models achieve the marginal coverage guarantee of $(1-\alpha_o)(1-\alpha_r)$ shown in Corollary 1. This observation highlights the high predictive uncertainty of existing SGG methods, as well as, shed light on possible distribution shifts in the train and test splits of SGG datasets that make computing well-calibration conformal quantiles very challenging under exchangeability assumptions. 

\subsection{Results of MLLM-based Post-Processing}
Table \ref{tab:llm_results} shows how the average set size and empirical coverage values of the whole triplet prediction sets are impacted by applying our proposed MLLM-based plausibility assessment method as a post-processing unit. It is observed that our proposed post-processing strategy is indeed effective in significantly shrinking the size of the original triplet prediction set constructed from the conformal prediction. Additionally, in most cases, the empirical coverage $Cov_T$ remains unaffected; only for SQUAT and MOTIFS-D a slight decrease in empirical coverage is obsevered. However, this small decrement in empirical coverage is tolerable given $> 50\%$ shrinkage in the sizes of their respective triplet prediction sets. Although the triplet prediction sets of BGNN have the highest empirical coverage, it must also be noted its prediction sets are also of the largest size since its predicate classifier is comparatively more uncertain than the other methods, as evident from Table \ref{tab:coverage_comparison}.

\begin{table}[t]
\centering
\caption{Comparison of cR@\textit{K} and  cmR@\textit{K} of the different SGG methods when PC-SGG is added. Performance of these methods for standard SGG based on R@\textit{K} and mR@\textit{K} are also shown.}
 \resizebox{1\linewidth}{!}{%
\begin{tabular}{lcccc}
\toprule
\textbf{Method} & \textbf{R@50} & \textbf{R@100} & \textbf{mR@50} & \textbf{mR@100} \\
\hline
MOTIFS~\cite{zellers2018neural} & 23.61 & 29.08 & 4.52 & 6.22 \\
MOTIFS-D~\cite{cui2019class} & 24.33 & 30.12 & 5.26 & 7.06 \\
VCTREE~\cite{tang2019learning} & 26.77 & 31.46 & 5.73 & 7.14 \\
SQUAT~\cite{jung2023devil} & 26.81 & 32.06 & 9.95 & 12.05 \\
BGNN~\cite{li2021bipartite} & 30.07 & 34.90 & 9.63 & 11.92 \\
\midrule
\midrule
\textbf{Method+PC-SGG} & \textbf{cR@50} & \textbf{cR@100} & \textbf{cmR@50} & \textbf{cmR@100} \\
\hline
MOTIFS~\cite{zellers2018neural} & 38.45 & 46.79 & 25.49 & 34.03 \\
MOTIFS-D~\cite{cui2019class} & 40.21 & 47.46 & 26.17 & 35.63 \\
VCTREE~\cite{tang2019learning} & 41.89 & 49.90 & 27.84 & 36.75 \\
SQUAT~\cite{jung2023devil} & 43.23 & 51.87 & 30.94 & 39.23 \\
BGNN~\cite{li2021bipartite} & \textbf{46.32} & \textbf{53.81} & \textbf{32.52} & \textbf{40.36} \\
\bottomrule
\end{tabular}
}
\label{tab:recall_comparison}
\end{table}

\begin{figure}
    \centering
    \includegraphics[width=0.9\linewidth]{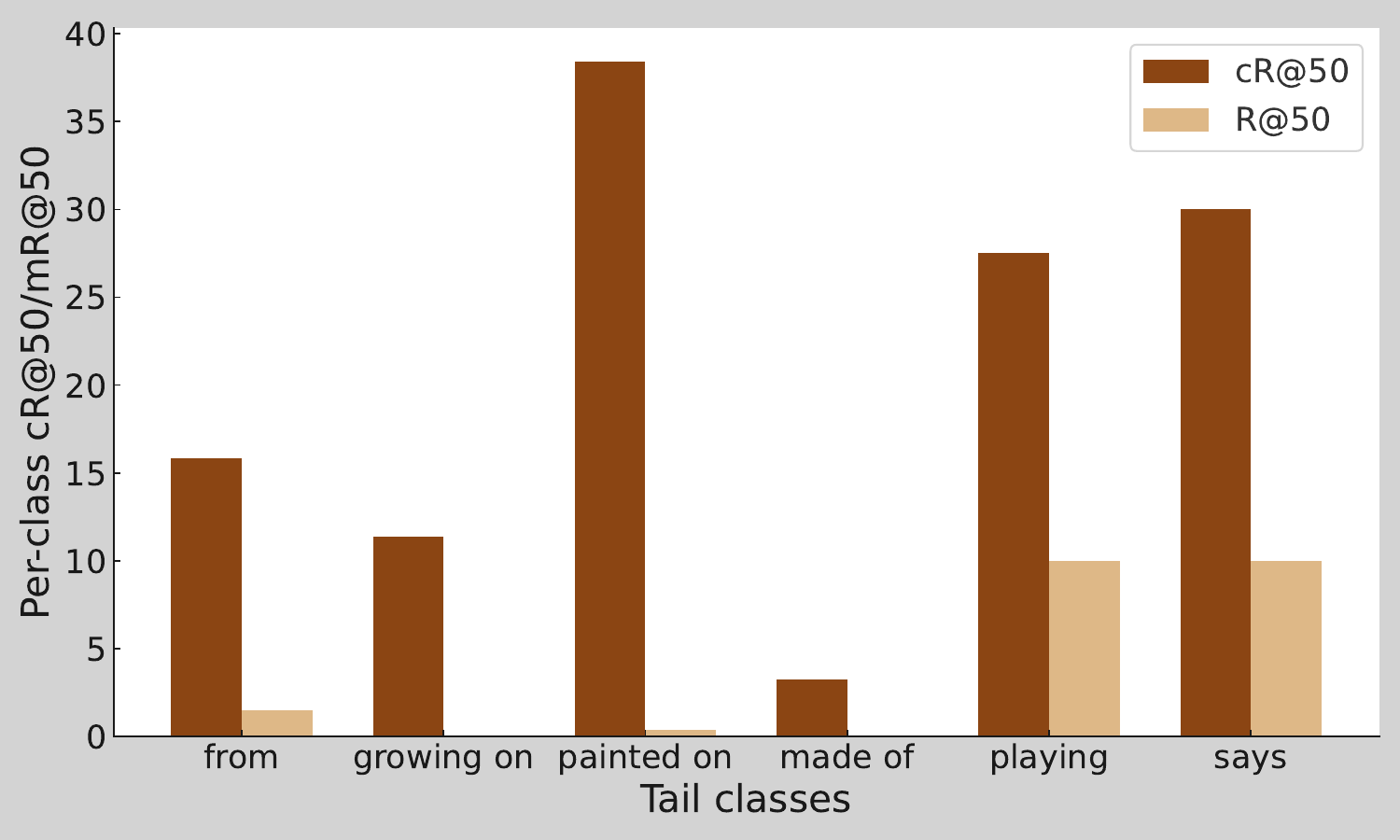}
    \caption{The R@50 and cR@50 for $5$ of tail classes of VG150, with the least number of samples. The results are for the BGNN model. It can be observed that the triplet prediction sets from BGNN+PC-SGG, significantly improve the performance of detecting the tail class predicates compared to the standalone model.}
    \label{fig:per_class_recall}
\end{figure}

\subsection{Comparsion of SGG Metrics after incorporating PC-SGG}

From Table \ref{tab:recall_comparison} one can observe that BGNN+PC-SGG achieves the highest cR@\textit{K} and cmR@\textit{K} values, showcasing the best \emph{conformal SGG} performance across all the SGG models used in this paper.
When PC-SGG is added, the recall-hit rate improves significantly since the set-valued triplet predictions provide a much greater chance of covering the actual ground truth than the standard SGG scenario, where each detected triplet has only one prediction instead of a set of predictions. As observed in Table \ref{tab:recall_comparison}, the recall-hit rate improves by an average of $15.70$ $\%$ points when comparing cR@$50$ with R@$50$.

Also, since PC-SGG is designed to achieve class-conditional coverage guarantees, adding it to any SGG method significantly improves performance on the tail classes. Across all SGG methods, we can observe that the class-conditional recall-hit rate improves by an average of $21.57$ $\%$ points when comparing cmR@$50$ with mR@$50$.
Note that the R@\textit{K} and mR@\textit{K} values of the standalone models differ from those reported in their respective papers since we retrain them on a new training split of VG150. Some qualitative results are provided in the supplementary.

\subsection{Analysis}

\begin{figure}[t]
    \centering
    \begin{subfigure}[t]{0.45\linewidth}
        \centering
        \includegraphics[width=\linewidth]{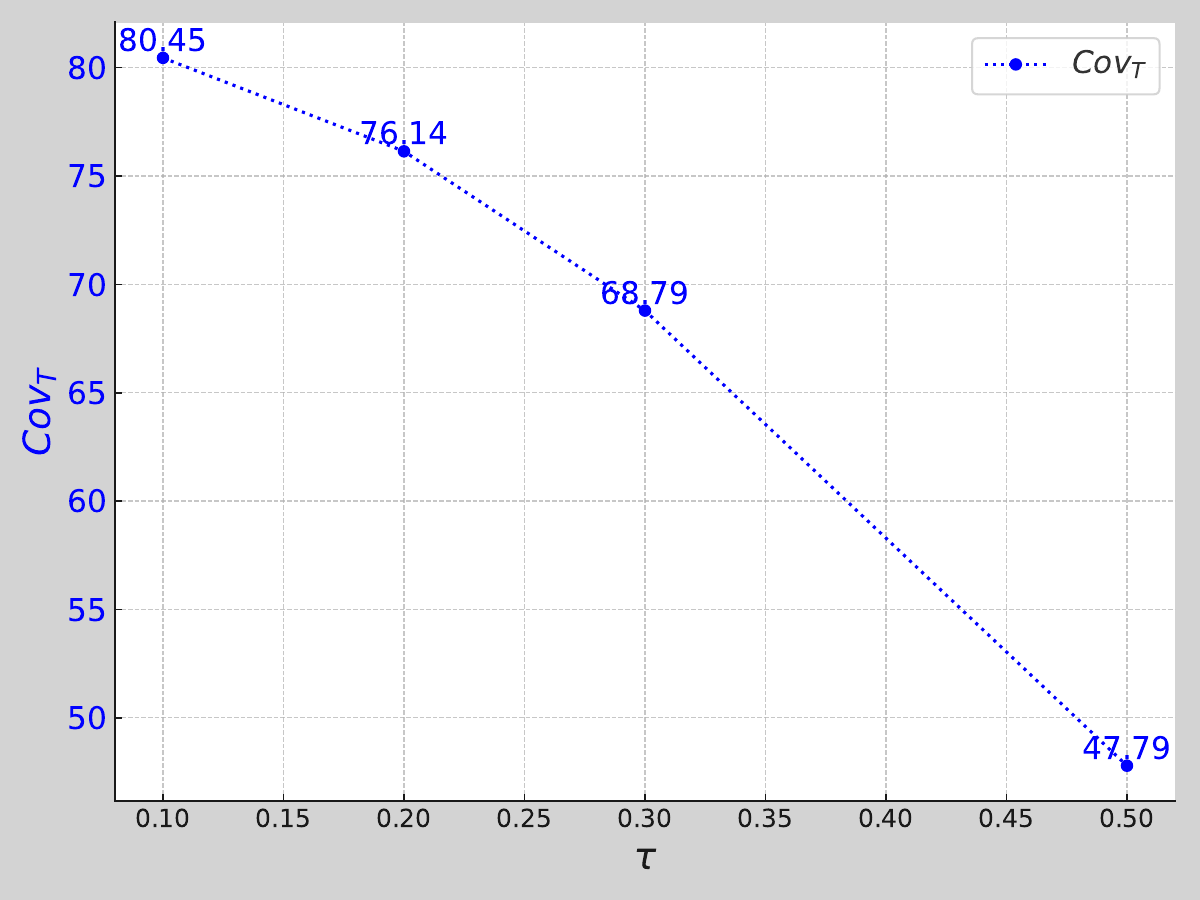} 
        \caption{$Cov_T$ vs $\tau$}
        \label{fig:cov_t_vs_tau}
    \end{subfigure}
    \hfill
    \begin{subfigure}[t]{0.45\linewidth}
        \centering
        \includegraphics[width=\linewidth]{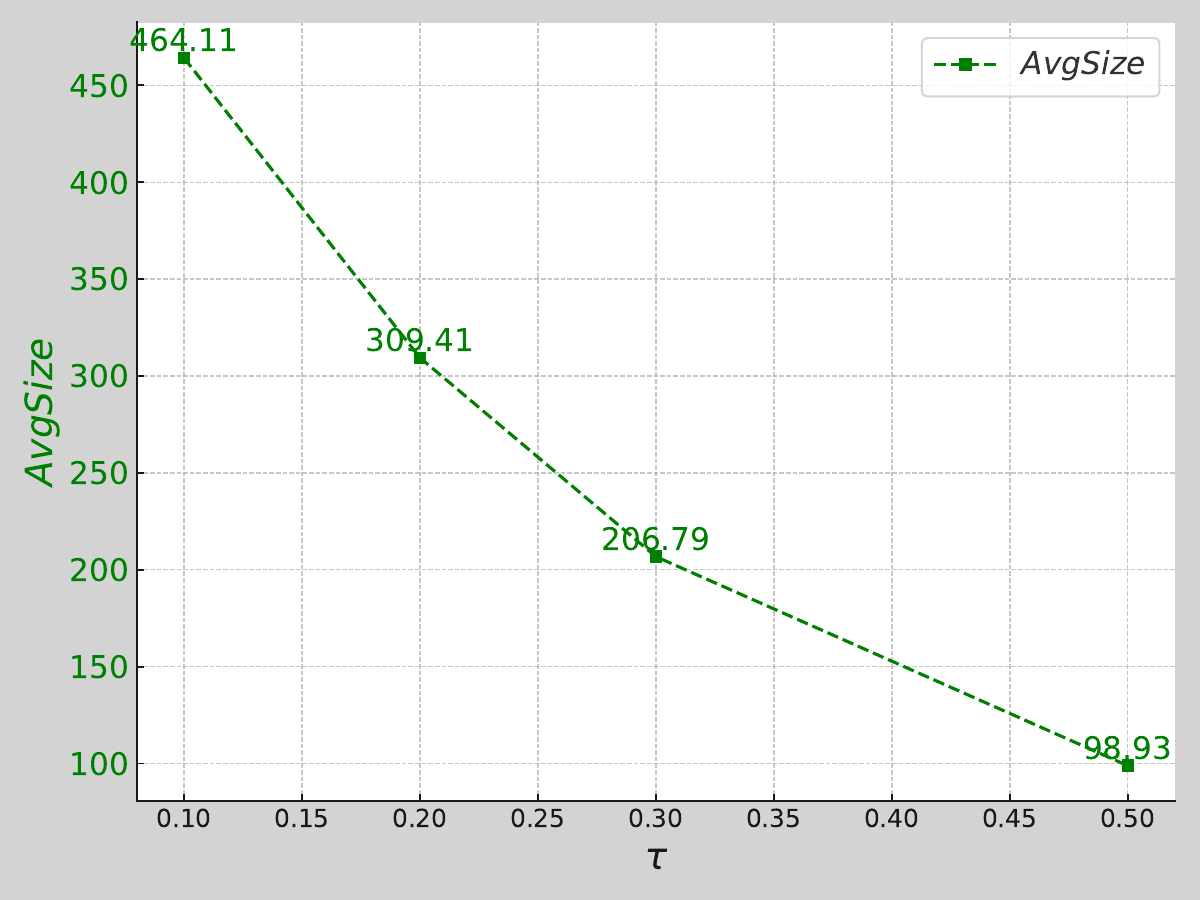} 
        \caption{$AvgSize$ vs $\tau$}
        \label{fig:avgsize_vs_tau}
    \end{subfigure}
    \caption{Comparison of $Cov_T$ and $AvgSize$ as functions of token threshold, $\tau$, for BGNN+PC-SGG.}
    \vspace{-1em}
    \label{fig:analysis1}
\end{figure}

We analyze different designs for the MLLM based post-processing strategy for the BGNN model's prediction sets.

\noindent\textbf{Impact of token threshold.} As mentioned in Sec \ref{sec:MLLM}, the choice $\tau$ significantly affects the empirical coverage, as well as, the size of the truncated triplet prediction sets. As observed in Fig \ref{fig:analysis1} varying $\tau$ between $\{0.1,0.2,0.3,0.5\}$ shows when $tau$, although the $AvgSize$ decreases, it comes at the cost of a drastic fall in empirical triplet coverage. This is because, in multiple scenarios, the tokens predicted by the MLLM with very high probabilities may map to plausible descriptions of the scene in the test image but may not be the actual ground truth scene description. Consequently, the truncated prediction sets omit the ground truth itself. Therefore, for all experiments, we choose $\tau=0.1$.

\begin{figure}[t]
    \centering
    \includegraphics[width=0.9\linewidth]{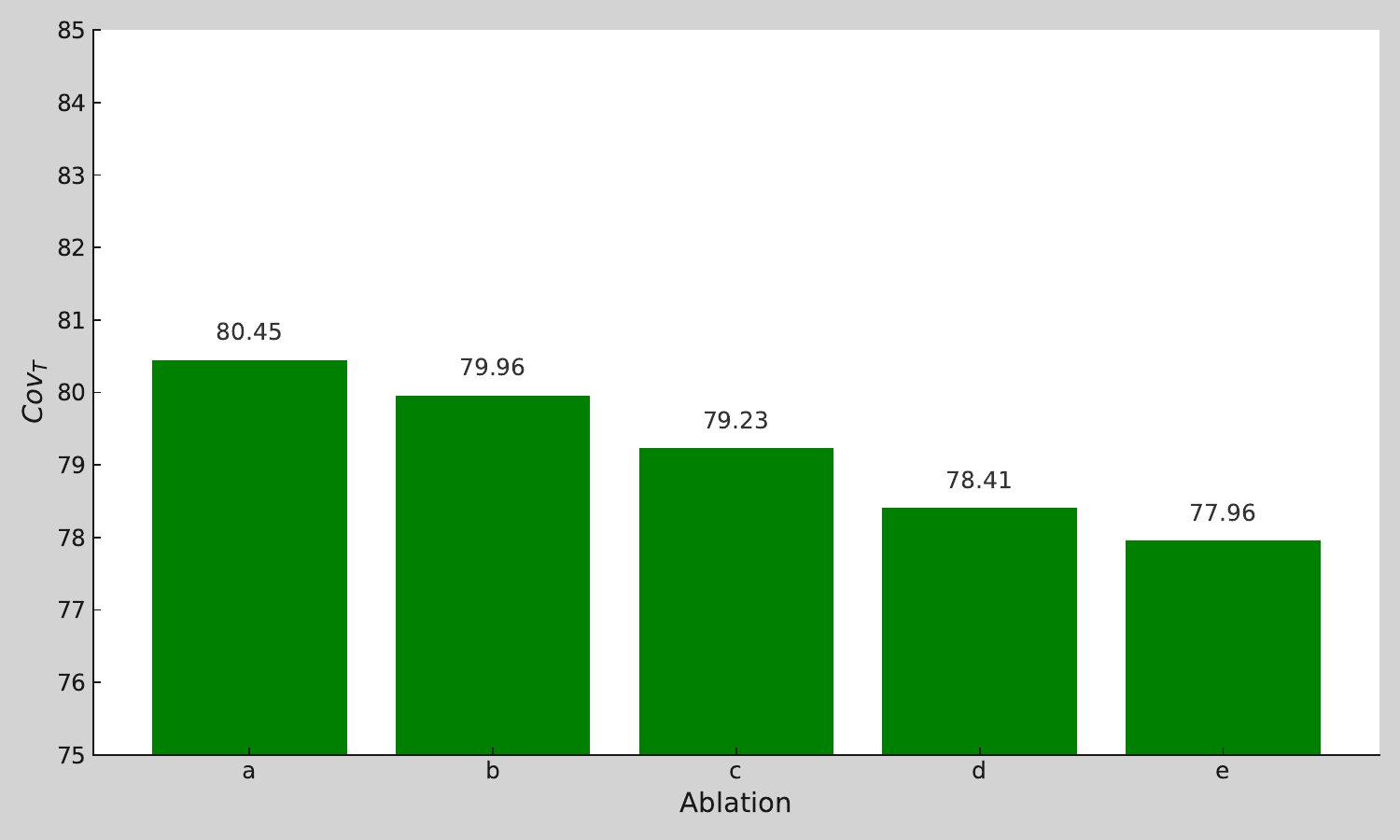}
    \caption{\textbf{Impact of different prompting strategies on $Cov_T$}, (a)Proposed prompting design, (b) No Image Cropping, (c) No System Prompt, (d) No Example prompt, (e) No System and example prompt. All results are obtained for the BGNN model.  }
    \vspace{-1em}
    \label{fig:prompting_cov}
\end{figure}

\noindent\textbf{Impact of image cropping.} From Fig \ref{fig:prompting_cov}, we can observe that if the test image is not cropped with the union feature of the subject and object bounding boxes, the empirical coverage drops. This is inherent because the visual information is not restricted to the triplet but involves other background information that distorts the MLLM's perception, resulting in the prediction set entries that have a low correlation with the cropped scene being incorrectly chosen to be part of the truncated prediction sets.

\noindent\textbf{Impact of in-context learning.} By comparing the empirical triplet coverage values of (a) with (c), (d), and (e) shown in Fig \ref{fig:prompting_cov}, we can observe that the System and Example Prompts are critical for the MLLM's understanding of the task at hand, as without the system prompt and the one-shot in-context example there is a significant drop in empirical coverage. In fact, we observe that without the in-context example, the token prediction space does not remain restricted to the MCQ choices, and random tokens are also predicted. More analysis is provided in the supplementary.
\section{Conclusion}
\label{sec:conc}
We present a novel post-hoc model agnostic method for quantifying the uncertainty in scene graph generation methods, such that their reliability can be gauged for safety-critical applications. To this end, we propose \textbf{P}lausibility ensured \textbf{C}onformal \textbf{SGG} or \textbf{PC-SGG}. PC-SGG is comprised of a novel conformal prediction framework specific to SGG that facilitates uncertainty quantification by providing prediction sets of the possible scene graphs in an image. To ensure the plausibility of the scene graphs in each prediction set, we also propose a novel MLLM-based plausibility assessment strategy as a post-processing unit of PC-SGG. We introduce an effective MCQA style prompting strategy for tasking an MLLM to assess the plausibility of the scene graphs and truncate the prediction sets. We show how PC-SGG quantifies the predictive uncertainty of numerous SGG methods. We also show that utilizing PC-SGG significantly improves SGG performance as the generation of multiple plausible scene graphs in an image improves the chances of capturing the ground-truth graph. In future work, we would extend the method to video scene graph generation.

\noindent\textbf{Acknowledgments.}This work was partially supported by NSF grants OAC-2411453, CCF-2008020, CNS-2312395, CMMI-2326309 and DURIP N000141812252.


{
    \small
    \bibliographystyle{ieeenat_fullname}
    \bibliography{main}
}

\clearpage
\newpage
\onecolumn



\section{Algorithmic and Mathematical Details}

\subsection{Algorithm for Conformal Scene Graph Generation}

The overall algorithm for our proposed SGG-specific CP is shown in Algorithm \ref{algo:class_conditional_cp}. The following details are important regarding the algorithm:
\\
\begin{itemize}
    \item The training and calibration sets are denoted as $\mathcal{D}_{tr}=\{I_j,G_j\}_{j=1}^{m}$ and $\mathcal{D}_{cal}=\{I_j,G_j\}_{j=m+1}^{m+n}$, where $I_j$ is an image and $G_j$ is its ground truth scene graph. Each $G$ is composed of triplets belonging to the object and predicate classes ($\mathcal{Y}_o,\mathcal{Y}_r$) as described in Sec 3.2. A test image is denoted as $I_{n+1}$.\\

    \item It must be noted that when referring to a sample, $X_j$, for a prediction set $\hat{\mathcal{C}}(X_j)$, we do not refer to the whole image. Specifically for the object prediction set $\hat{\mathcal{C}}_o(X_i)$, $X_j$ refers to a single RoI describing a detected object. For the predicate prediction set $\hat{\mathcal{C}}_r(X_j)$, the sample $X_j$ refers to a pair of detected objects, as information about the pair including the union box of its two objects is included in the sample. In some cases like in Algorithm \ref{algo:class_conditional_cp} we specifically distinguish the samples for the object and predicate classes by denoting them as $X_j^o$ and $X_j^r$ respectively (the meaning remains the same). \\

    \item The assumption of exchangeability i.e. $\mathcal{D}_{cal}\cup {(I_{n+1},G_{n+1})}$ also
    implies exchangeability holds for any subsets of $D_{cal}$, such as the considered partitions $\mathcal{D}_{cal,y}^o \subset \mathcal{D}_{cal}$, and $\mathcal{D}_{cal,y}^r \subset \mathcal{D}_{cal}$. \\

    \item Algorithm \ref{algo:class_conditional_cp} can sometimes result in null sets for the objects and predicates, however, such an event occurred extremely rarely during our experiments. In such cases, we follow common practice~\cite{timans2024adaptive} and choose the class with the highest softmax probability value i.e. $\underset{y\in\mathcal{Y}}{max} \ \pi_y$, as the prediction set. As such a singleton prediction set is constructed in such cases.
    
 \end{itemize}
 
 \begin{algorithm}[t]
\caption{Conformal Scene Graph Generation}
\label{algo:class_conditional_cp}
\begin{algorithmic}[1]
    
    \State \textbf{Input:} Training Set: $\mathcal{D}_{tr}$, Calibration Set: $\mathcal{D}_{cal}$, Object Miscoverage Rate: $\alpha_o$, Predicate Miscoverage Rate: $\alpha_r$, 
    
    \noindent Test Image: $I_{n+1}$. 
    
    \State \textbf{Output:} Object Prediction Set: $\hat{\mathcal{C}}_o(X_{n+1}^o)$, Predicate Prediction Set: $\hat{\mathcal{C}}_r(X_{n+1}^r)$, Triplet Prediction Set: $\hat{\mathcal{C}}_t(X_{n+1}^r)$.\\
    \algrule 
    
    \State Fit an SGG model, $\phi$, on the training set $\mathcal{D}_{train}$.

    \State \textbf{Calibration Procedure:} 
    
    \State Assume the calibration set $\mathcal{D}_{cal}$ which is comprised of images has the following subsets, 
    
    $\mathcal{D}_{cal, y}^o = \{ (X_i^o,Y_i) : Y_i = y_i^o \}, \forall y_i^o \in \mathcal{Y}_o$. 
    
    $\mathcal{D}_{cal, y}^r = \{ (X_i^r,Y_i) : Y_i = y_i^r \}, \forall y_i^r \in \mathcal{Y}_r$. 
    
    \noindent where, $\mathcal{D}_{cal, y}^o,\mathcal{D}_{cal, y}^r \in \mathcal{D}_{cal} $ define a classwise calibration subset specific to the object and predicate classes in the images of $\mathcal{D}_{cal}$, and $X_i^o \in \mathcal{D}_{cal, y}^o$ refers to the object classification specific sample defined by an RoI in the image, $X_i^r \in \mathcal{D}_{cal, y}^o$ refers to the predicate classification specific sample defined by a pair of objects~\cite{zellers2018neural,jung2023devil,li2021bipartite}. 
    
    \State Define a nonconformity function 
    
    $\mathcal{A}:\mathcal{X} \times \mathcal{Y} \rightarrow [0,1], \, (\hat{f}(X), y) \mapsto 1 - \hat{\pi}_{y}(X)$ 

    \noindent $\hat{f}$ is any classifier, $\hat{\pi}_{y}(X)$ is estimated true class probability. Therefore, the complement of $\hat{\pi}_{y}(X)$    encodes a notion of dissimilarity (nonconformity) between the predicted and true class probabilities.

    \State Define list of object and predicate class quantiles $Q_{o}$ and $Q_r$.

    \State Match pair of detected objects with ground-truth pair of objects using Eq 3. 

    \State \textbf{Begin for each} $y_i^o \in \mathcal{Y}_o$ \textbf{and} $y_i^r \in \mathcal{Y}_r$:
    
    \State Apply $\mathcal{A}$ to $\mathcal{D}_{cal, y}^o$ to obtain a set of scores 
    
    $S_{y_i^o} = \{\mathcal{A}(f_o(X_i),y_i^o)\}_{i=1}^{n_{y_i^o}} = \{s_i^o\}_{i=1}^{n_{y_i^o}}$.  where $f_o$ is the object classifier within $\phi$

    \State Apply $\mathcal{A}$ to $\mathcal{D}_{cal, y}^r$ to obtain a set of scores 
    
    $S_{y_i^r} = \{\mathcal{A}(f_r(X_i),y_i^r)\}_{i=1}^{n_{y_i^r}} = \{s_i^r\}_{i=1}^{n_{y_i^r}}$.  where $f_r$ is the predicate classifier within $\phi$
    
    \State Compute a conformal quantiles $\hat{q}_{y_i^o}$ and $\hat{q}_{y_i^r}$, defined as, 
    
    $\hat{q}_{y_i^o} = \lceil (n_{y_i^o}+1)(1-\alpha_o)/n_{y_i^o} \rceil$-th empirical quantile of $S_{y_i^o}$.

    $\hat{q}_{y_i^r} =\lceil (n_{y_i^r}+1)(1-\alpha_r)/n_{y_i^r} \rceil$-th empirical quantile of $S_{y_i^r}$.


    \State Add object class quantile to the set: $Q_{o} = Q_o \cup \{ \hat{q}_{y_i^o} \}$, and Add predicate class quantile to the set: $Q_{r} = Q_r \cup \{ \hat{q}_{y_i^r} \}$.

    \State \textbf{End for}
    \State \textbf{End procedure}
    \State \textbf{Conformal Inference Procedure:}
    
    \State For a new test Image $I_{n+1}$ comprised of objects depicted by $(X_{n+1}^o,Y_{n+1}^o)$, and predicates depicted by $(X_{n+1}^r,Y_{n+1}^r)$,  valid prediction sets for $X_{n+1}^o$ and $X_{n+1}^r$ are constructed as,

    $\hat{\mathcal{C}}_o(X_{n+1}^o) = \{y_k^o \in \mathcal{Y}_o : \hat{\pi}_{y_{k}^o} \geq 1- \hat{q}_{y_k^o}\}$ 
    
    $\hat{\mathcal{C}}_r(X_{n+1}^r) = \{y_k^r \in \mathcal{Y}_r : \hat{\pi}_{y_{k}^r} \geq 1- \hat{q}_{y_k^r}\}$

    \noindent where $\hat{q}_{y_k^r} \in Q_r$ and $\hat{q}_{y_k^o} \in Q_o$. The validity of the sets refers to satisfying class-conditional coverage guarantee (Eq 2) with probability $(1-\alpha_o)$ for each object class, and $(1-\alpha_r)$ for each predicate class. (Proof: Sadinle \emph{et al.} ~\cite{m.sadinle2019a})

    \State Combinatorially combine $\hat{\mathcal{C}}_o(X_{n+1}^o)$ and $\hat{\mathcal{C}}_r(X_{n+1}^r)$ to construct a triplet prediction $\hat{\mathcal{C}}_t(X_{n+1}^r)$.
    
    \State \textbf{End procedure}
\end{algorithmic}
\end{algorithm}

\subsection{Nominal Coverage Guarantee of Triplet prediction Sets}

\setcounter{theorem}{0}
\begin{theorem}
    Given the ground truth class of the $k^{th}$ triplet is denoted as $y_k^t=[y_{k}^s,y_{k}^r,y_{k}^o] \in \mathbb{R}^3$ where $y_{k}^s,y_{k}^o \in \mathcal{Y}_o$ and $y_{k}^r\in \mathcal{Y}_r$, the triplet coverage guarantee is given as $P(y_k^t \in \hat{\mathcal{C}}_t(X_{n+1}^r)) = P(Y_{n+1}^o \in \hat{\mathcal{C}}_o(X_{n+1}^o) \mid Y_{n+1}^o = y_i^o) \cdot P(Y_{n+1}^r \in \hat{\mathcal{C}}_o(X_{n+1}^r) \mid Y_{n+1}^r = y_m^r) \ \forall \ y_{k}^s\in \mathcal{Y}_o, y_{k}^o \in \mathcal{Y}_o, y_{k}^r \in \mathcal{Y}_r$.
\end{theorem}


\begin{proof}
    Assume the ground truth of the $k^{th}$ triplet in an image is denoted as $y_k^t=[y_k^s,y_k^r,y_k^o] \in \mathbb{R}^3$ where $y_k^s,y_k^o \in \mathcal{Y}_o$ are the ground-truth classes of the subject-object pair and $y_k^r \in \mathcal{Y}_r$ is the ground-truth class of the predicate. Since $y_k^s,y_k^o \in \mathcal{Y}_o$, for clarity let's denote $y_k^s = y_i^o \ | \ i \in  [1,K_o]$, $y_k^o = y_j^o \ | \ j \in  [1,K_o] \ \& \ j \neq i$, and $y_k^r = y_m^r \ | \ m \in [1,K_r]$. Now assuming the prediction sets for the subject and object are given as  $\hat{\mathcal{C}}_o(X_{n+1}^o)$, $\hat{\mathcal{C}}_o(X_{n+2}^o)$, and the prediction set of the predicate is given as $\hat{\mathcal{C}}_r(X_{n+1}^r)$, the nominal coverage guarantee of a triplet prediction set is described as,

    \begin{equation}
    \scalebox{0.94}{$
        P(y_k^t \in \hat{\mathcal{C}}_t(X_{n+1}^r)) = P(Y_{n+1}^o \in \hat{\mathcal{C}}_o(X_{n+1}^o) \land 
        Y_{n+2}^o \in \hat{\mathcal{C}}_o(X_{n+2}^o) \land 
        Y_{n+1}^r \in \hat{\mathcal{C}}_r(X_{n+1}^r)|Y_{n+1}^o=y_i^o, Y_{n+2}^o=y_j^o,Y_{n+1}^r=y_m^r )
    $}
    \end{equation}

    \noindent We observe that the guarantees of the object and predicate prediction are controlled by distinct conformal procedures on the calibration data and, as such, are conditionally independent. Therefore,

    \begin{equation}
    \scalebox{0.94}{$
        P(y_k^t \in \hat{\mathcal{C}}_t(X_{n+1}^r)) =  P(Y_{n+1}^o \in \hat{\mathcal{C}}_o(X_{n+1}^o) \land 
        Y_{n+2}^o \in \hat{\mathcal{C}}_o(X_{n+2}^o) | Y_{n+1}^o=y_i^o, Y_{n+2}^o=y_j^o) \cdot P(Y_{n+1}^r \in \hat{\mathcal{C}}_o(X_{n+1}^r)|Y_{n+1}^r=y_m^r)
     $}
    \end{equation}

    \noindent Additionally, there is no separate subject and object detection, as all objects in an image are detected once and then combinatorially combined to form subject-object pairs. Therefore, the class-conditional coverage guarantee of the subject is contained in the class-conditional coverage guarantee of the object. Hence,

    \begin{flalign}
    & P(Y_{n+1}^o \in \hat{\mathcal{C}}_o(X_{n+1}^o) \land 
        Y_{n+2}^o \in \hat{\mathcal{C}}_o(X_{n+2}^o) | Y_{n+1}^o=y_i^o, Y_{n+2}^o=y_j^o) =  P(Y_{n+1}^o \in \hat{\mathcal{C}}_o(X_{n+1}^o) | Y_{n+1}^o=y_i^o) &\label{eq1} \\
    &\implies P(y_k^t \in \hat{\mathcal{C}}_t(X_{n+1}^r)) = P(Y_{n+1}^o \in \hat{\mathcal{C}}_o(X_{n+1}^o) \mid Y_{n+1}^o = y_i^o) \cdot P(Y_{n+1}^r \in \hat{\mathcal{C}}_o(X_{n+1}^r) \mid Y_{n+1}^r = y_m^r) &\label{eq2} 
\end{flalign}
\end{proof}

\vspace{2em}
\setcounter{corollary}{0} 
\begin{corollary}
    Following Theorem 1, $P(y_k^t \in \hat{\mathcal{C}}_t(X_{n+1}^r)) \geq (1-\alpha_o)(1-\alpha_r), \ \forall \ y_{k}^s\in \mathcal{Y}_o, y_{k}^o \in \mathcal{Y}_o, y_{k}^r \in \mathcal{Y}_r$.
\end{corollary}
\begin{proof}
This follows trivially from Theorem 1. Since
\begin{flalign}
& P(Y_{n+1}^o \in \hat{\mathcal{C}}_o(X_{n+1}^o) \mid Y_{n+1}^o = y_i^o) \geq (1-\alpha_o) \quad and \quad P(Y_{n+1}^r \in \hat{\mathcal{C}}_o(X_{n+1}^r) \mid Y_{n+1}^r = y_m^r) \geq (1-\alpha_r) \\
&\implies P(y_k^t \in \hat{\mathcal{C}}_t(X_{n+1}^r)) \geq (1-\alpha_o) \cdot (1-\alpha_r) &\label{eq3}
\end{flalign}
\end{proof}

It must be pointed out that the nominal coverage guarantee in Eq \ref{eq3} is the intended coverage goal, based on which calibration is conducted (under assumptions of exchangeability). However, in practice, the empirical coverage may not always reach coverage guarantees, owing to the high predictive uncertainty of the underlying model as well as, unquantified/subtle distribution shifts between the train/calibration and test data~\cite{romano2020classification,angelopoulos2021uncertainty,m.sadinle2019a,timans2024adaptive}. We observe this in our empirical results (Table 1). Additionally, from an empirical standpoint, the results in Table 1, of the main paper, validate Eq \ref{eq2} in the sense that the empirical coverage of the triplet prediction set is approximately close to the product of the empirical coverages of the object and predicate prediction sets.

\section{Additional Implementation Details}

\subsection{Computation of Conformal Prediction Metrics }

End-to-end SGG or SGDET entails the localization and classification of all objects in a scene along with classifying their pairwise predicates. As such, when computing the CP metrics on the inference data, we need to match the predicted bounding boxes with the ground-truth ones in an image. We do so by following the same pairwise greedy matching strategy shown in Eq 3. Finally, the CP metrics are computed over the matched predictions only as is standard practice in CP-based localization studies~\cite{li2022towards,timans2024adaptive}.

\begin{figure}[t] 
    \centering
    
    \begin{subfigure}{0.32\textwidth}
        \includegraphics[width=\linewidth]{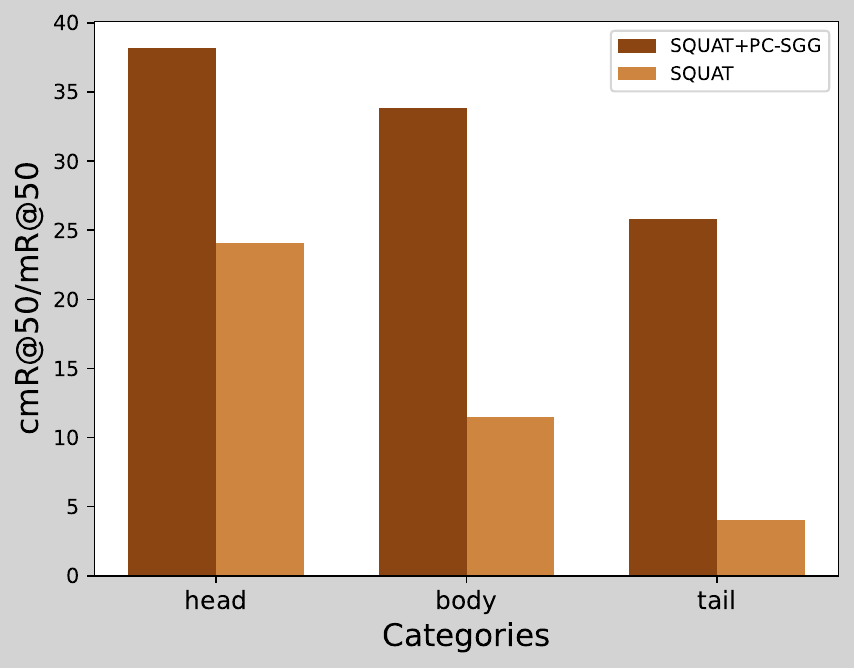}
        \caption{SQUAT}
        \label{fig:squat}
    \end{subfigure}
    \hfill
    \begin{subfigure}{0.32\textwidth}
        \includegraphics[width=\linewidth]{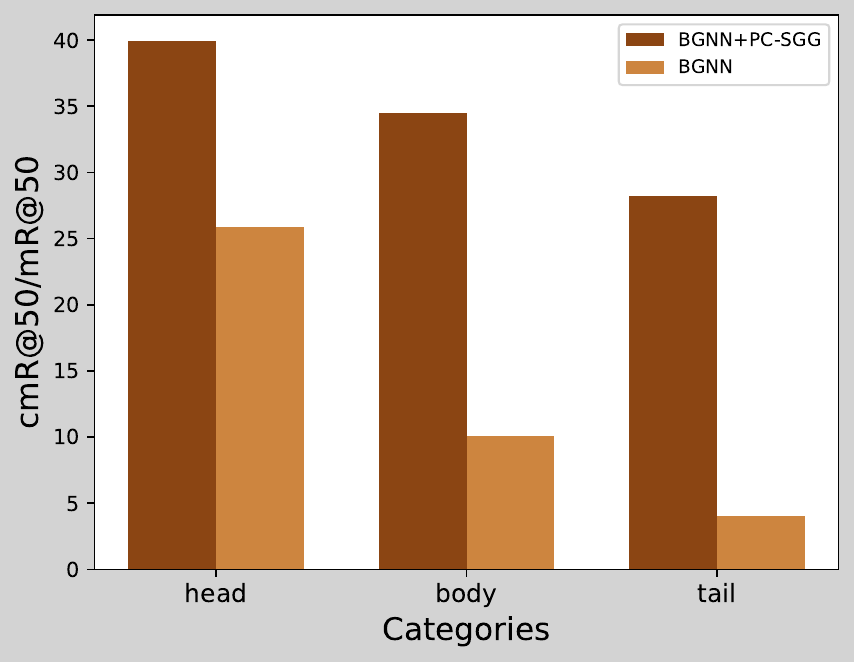}
        \caption{BGNN}
        \label{fig:bgnn}
    \end{subfigure}
    \hfill
    \begin{subfigure}{0.32\textwidth}
        \includegraphics[width=\linewidth]{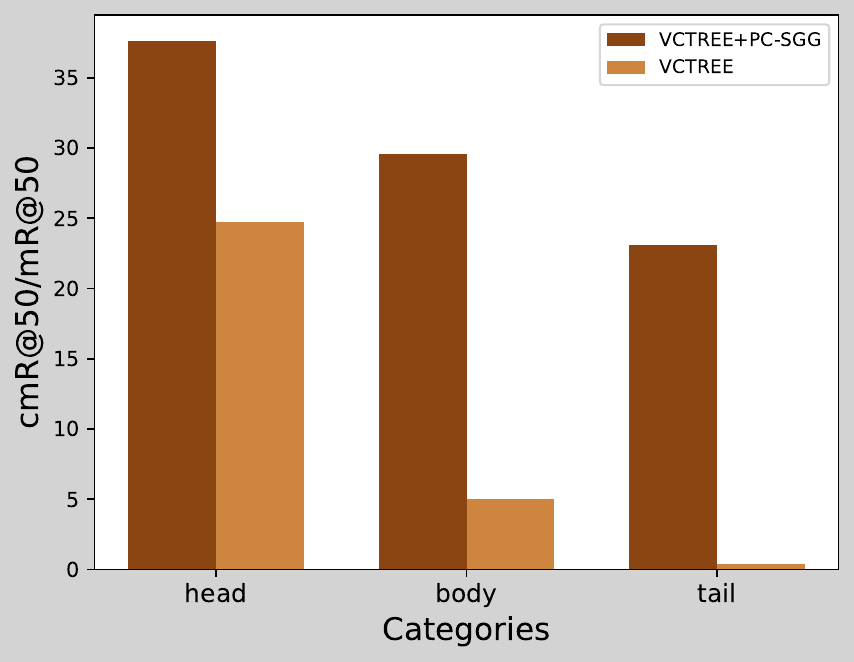}
        \caption{VCTREE}
        \label{fig:vctree}
    \end{subfigure}
    \hfill

    \vspace{0.5cm} 
    \begin{subfigure}{0.32\textwidth}
        \includegraphics[width=\linewidth]{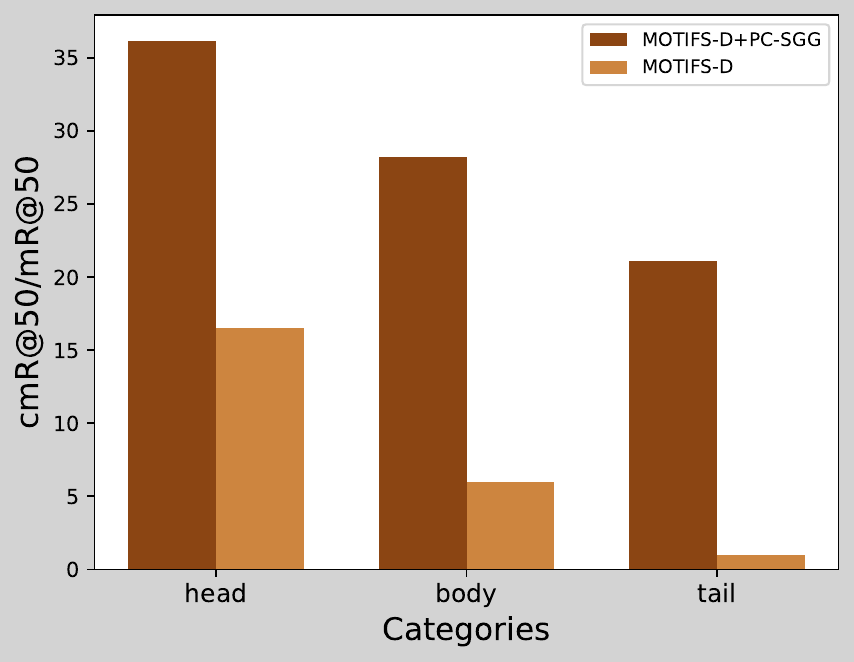}
        \caption{MOTIFS-D}
        \label{fig:motifs-d}
    \end{subfigure}
    \hspace{0.5em}
    \begin{subfigure}{0.32\textwidth}
        \includegraphics[width=\linewidth]{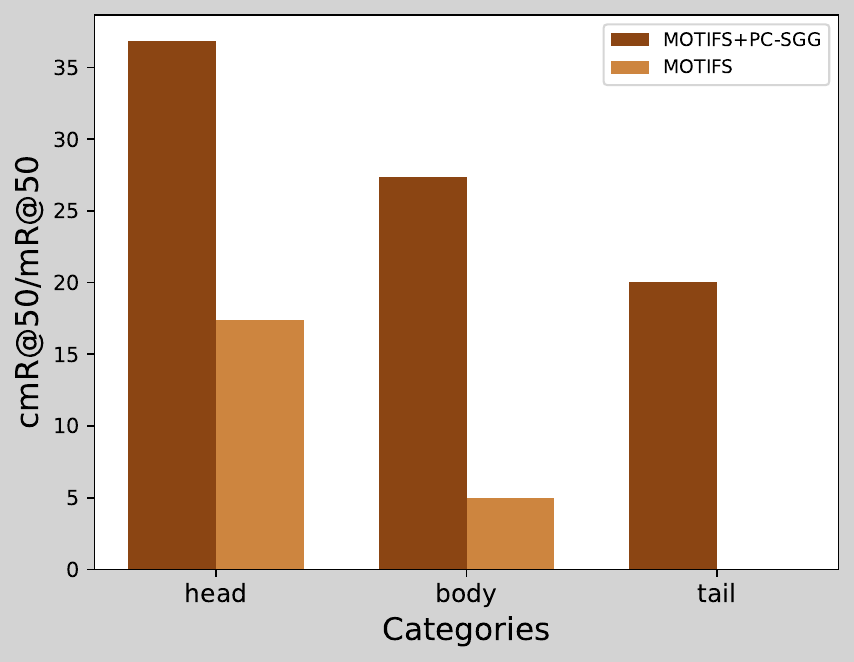}
        \caption{MOTIFS}
        \label{fig:motifs}
    \end{subfigure}
    \hfill


    \caption{Improvement of recall-hit rate across head, body, and tail categories of VG150~\cite{krishna2017visual}, with the incorporation of PC-SGG. The darker shade is the cmr@$50$ value and the lighter one is the mR@$50$ value.}
    \label{fig:comparison_hbt}
\end{figure}

\subsection{Adaptation of Scene Graph Generation Metrics for Prediction Sets}

For SGG the Recall@$K$ (R@$K$) and mean-Recall@$K$ (mR@$K$) are standard evaluation metrics~\cite{tang2019learning}. Formally if $K$ predicted triplets $\{\hat{t}_i\}_{i=1}^K$, are matched to a ground-truth triplet defined by the class $y_i$, then R@$K$ is computed as,
\begin{equation}
    R@K = \frac{1}{N} \sum_{i=1}^N \left( \bigvee_{j=1}^K \mathds{1}[y_i = \hat{t}_j] \right)
\end{equation}
where, $\bigvee$ is the logical OR operation, $\mathds{1}[\cdot]$ is the indicator function, $N$ is the total number of ground-truth triplets, and $\hat{t}_j,y_j \in \mathbb{R}^3$. However, R@$K$ is not designed for prediction sets and so to accommodate triplet prediction sets we propose coverage-Recall@$K$ (cR@$K$) which is computed as follows,
\begin{equation}
    cR@K = \frac{1}{N} \sum_{i=1}^N \left( \bigvee_{j=1}^K \mathds{1}[y_i \in \hat{\mathcal{C}}_{t,j}(X_{i}^r)] \right)
\end{equation}
where $\hat{\mathcal{C}}_{t,j}(X_{i}^r)$ is the triplet prediction set of the $j^{th}$ predicted triplet. Therefore, the only difference between R@$K$ and cR@$K$ is that the \emph{equality} ($=$) operation is replaced with the \emph{belongs to} ($\in$) operation. As such cR@$K$ is equivalent to R@$K$ when predictions are not in the form of a single triplet but in the form of prediction sets. The coverage-mean-Recall@$K$ (cmR@$K$) metric is designed similarly and is equivalent to the mR@$K$ metric.

\section{Additional Empirical Results}




\begin{table}[h]
\centering
\caption{\textbf{Impact of number of options in the MCQA prompt.} Results are shown for the BGNN model. The number of options here refers to the number of prediction set entries. The `no valid option' choice is not counted.}
\begin{tabular}{ccc}
\hline
\textbf{Number of MCQA Options}                   & $Cov_T\uparrow$ & $AvgSize\downarrow$ \\ \hline
Original \\ (w/o MLLM post-processing) & 80.45           & 971.69              \\ \hline
3                                   & \textbf{80.45}           &   539.72           \\
5                                   & \textbf{80.45}           & 464.11              \\
10                                  & 69.38         & \textbf{291.47}             \\ \hline
\end{tabular}
\label{tab:num_options}
\end{table}

\subsection{Per Class Performance Improvement}

Incorporating PC-SGG with any existing SGG method significantly improves the recall-hit rate for every class in the scene graph dataset. this is evident from Fig 4 in the main paper where we show the massive performance improvement over some of the tail classes of VG150~\cite{krishna2017visual} for the BGNN method. In Fig \ref{fig:comparison_hbt} we further show the aggregate improvement of recall-hit rate across the HEAD, BODY, and TAIL classes of VG150 when PC-SGG is added to any of the SGG methods used in this paper. Specifically for MOTIFS and VCTREE, which had negligible or zero mR@$50$ values for the TAIL classes, incorporating PC-SGG is significantly beneficial as the generated triplet prediction sets cover most of the TAIL classes.

\subsection{More Analysis}
\subsubsection{Impact of different prompting strategies on average set size of triplet prediction sets}

\begin{figure}
    \centering
    \includegraphics[width=0.5\linewidth]{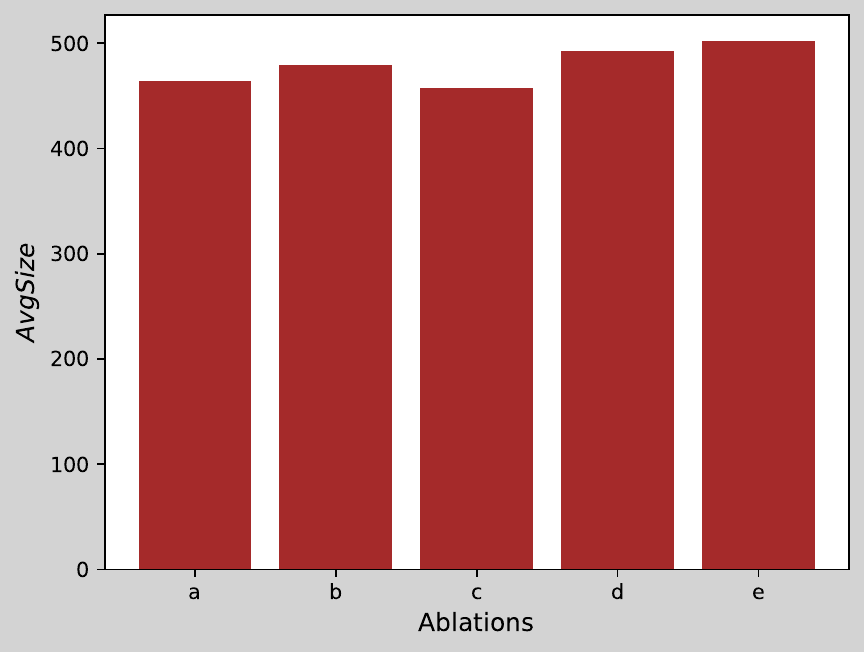}
    \caption{\textbf{Impact of different prompting strategies on $AvgSize$ of triplet prediction sets.} (a)Proposed prompting design, (b) No Image Cropping, (c) No System Prompt, (d) No Example prompt, (e) No System and Example prompt. All results are obtained for the BGNN model.  }
    \label{fig:prompting_avg_size}
\end{figure}


In Fig 6 of the main paper, we showed how incontext learning style prompting strategy benefits the MLLM-based plausibility assessment task, of truncating triplet prediction sets without impacting their original empirical coverage. 
Fig \ref{fig:prompting_avg_size} shows how changing our proposed prompting strategy impacts the average set size of the truncated prediction sets. We can observe the figure, that in general when the example prompt is not provided as a one-shot support example~\cite{brown2020language}, the $AvgSize$ increases, thus highlighting its importance for our task. 
On the other hand, the impact of not using image cropping or the system prompt is relatively small on the empirical set size of the truncated sets. However, given their impact on the $Cov_T$ values (Fig 6), it can be concluded that utilizing our full prompting strategy provides the most optimal results.

\subsubsection{Impact of number of options in the MCQA prompt }

We choose at max $5$ entries from a triplet prediction set, which along with the `no valid option' choice make a total of $6$ options in the MCQA prompt to the MLLM. We observe empirically that $5$ options give the optimal performance. This can be validated from Table \ref{tab:num_options}, which shows that increasing the number of options adversely affects the coverage while decreasing the number adversely affects the average set size.

This phenomenon occurs because increasing the number of options confuses the MLLM, causing it to hallucinate tokens that indicate all choices are equally plausible. This results in near-uniform likelihood values across all tokens (where each token represents an option). Consequently, the MLLM behaves like a naive or random guesser, selecting entries that are implausible and often not the ground truth. This significantly impacts empirical coverage. Conversely, limiting the number of options preserves the original empirical coverage but restricts the truncation of the prediction set. This happens because, in the task of plausibility assessment, the MLLM inherently compares the provided options. When the token space is reduced, the comparison is constrained, leading to the inclusion of more entries in the set. Thus, the optimal number of options for the MCQA prompt is $5$.

\subsection{Qualitative Visualization}

\begin{figure}
    \centering
    \includegraphics[width=1\linewidth]{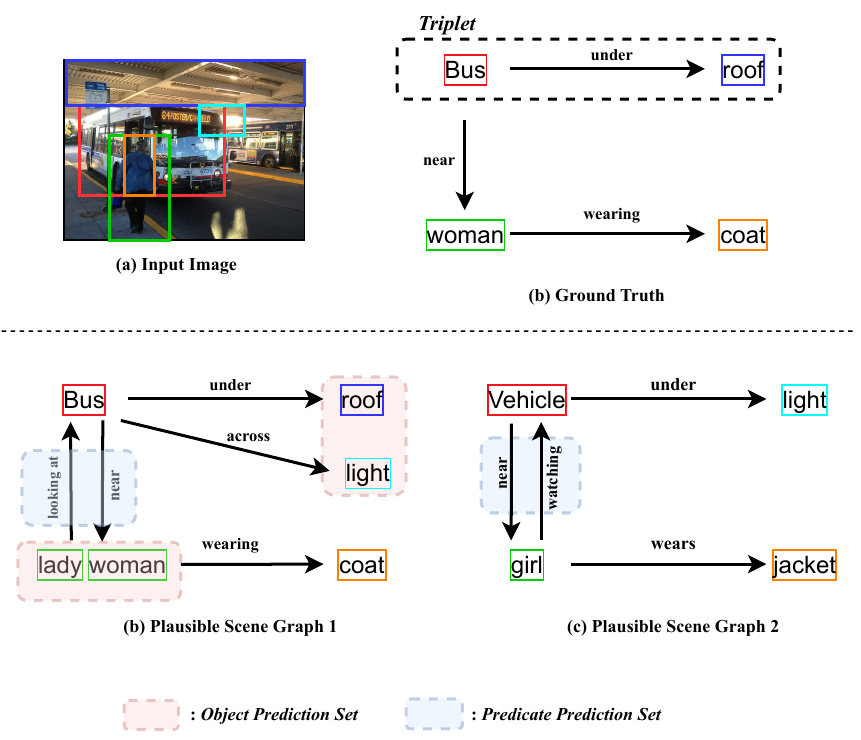}
    \caption{Qualitative results on a test image with BGNN+PC-SGG. The prediction sets obtained via PC-SGG facilitate the generation of multiple plausible scene graphs.}
    \label{fig:qualitative}
\end{figure}

Fig \ref{fig:qualitative} shows some plausible scene graphs generated by the BGNN+PC-SGG method. The importance of each plausible scene graph will depend on the downstream application and must be determined by domain expertise.

\newpage

\end{document}